%% file: arxiv.tex
\documentclass[letterpaper]{article} 
\usepackage{aaai2026}  
\usepackage{times}  
\usepackage{helvet}  
\usepackage{courier}  
\usepackage[hyphens]{url}  
\usepackage{graphicx} 
\usepackage{natbib}  
\usepackage{caption} 
\usepackage{xcolor}

\usepackage{algorithm}
\usepackage{algorithmic}

\usepackage{amssymb}
\usepackage{amsmath}
\usepackage{multirow}
\usepackage{multicol}
\usepackage{booktabs}
\usepackage{array}
\usepackage{makecell}
\usepackage{tabularx}
\usepackage{placeins}

\usepackage{newfloat}
\usepackage{listings}
\DeclareCaptionStyle{ruled}{labelfont=normalfont,labelsep=colon,strut=off} 
\floatstyle{ruled}
\newfloat{listing}{tb}{lst}{}
\floatname{listing}{Listing}
\title{AnoStyler: Text-Driven Localized Anomaly Generation\\ 
via Lightweight Style Transfer}

\author{
    Yulim So, Seokho Kang\thanks{Corresponding author.}\\
}
\affiliations{
    Sungkyunkwan University\\
    you0715@skku.edu, s.kang@skku.edu
}

\usepackage{bibentry}

\begin{document}
\maketitle


\begin{abstract}
Anomaly generation has been widely explored to address the scarcity of anomaly images in real-world data. However, existing methods typically suffer from at least one of the following limitations, hindering their practical deployment: (1) lack of visual realism in generated anomalies; (2) dependence on large amounts of real images; and (3) use of memory-intensive, heavyweight model architectures. 
To overcome these limitations, we propose \textit{AnoStyler}, a lightweight yet effective method that frames zero-shot anomaly generation as text-guided style transfer. Given a single normal image along with its category label and expected defect type, an anomaly mask indicating the localized anomaly regions and two-class text prompts representing the normal and anomaly states are generated using generalizable category-agnostic procedures. A lightweight U-Net model trained with CLIP-based loss functions is used to stylize the normal image into a visually realistic anomaly image, where anomalies are localized by the anomaly mask and semantically aligned with the text prompts. Extensive experiments on the MVTec-AD and VisA datasets show that \textit{AnoStyler} outperforms existing anomaly generation methods in generating high-quality and diverse anomaly images. Furthermore, using these generated anomalies helps enhance anomaly detection performance.
\end{abstract}

\begin{links}
    \link{Code}{https://github.com/yulimso/AnoStyler}
\end{links}

\section{Introduction}

\begin{figure}[!t]
\centering
\includegraphics[width=\linewidth]{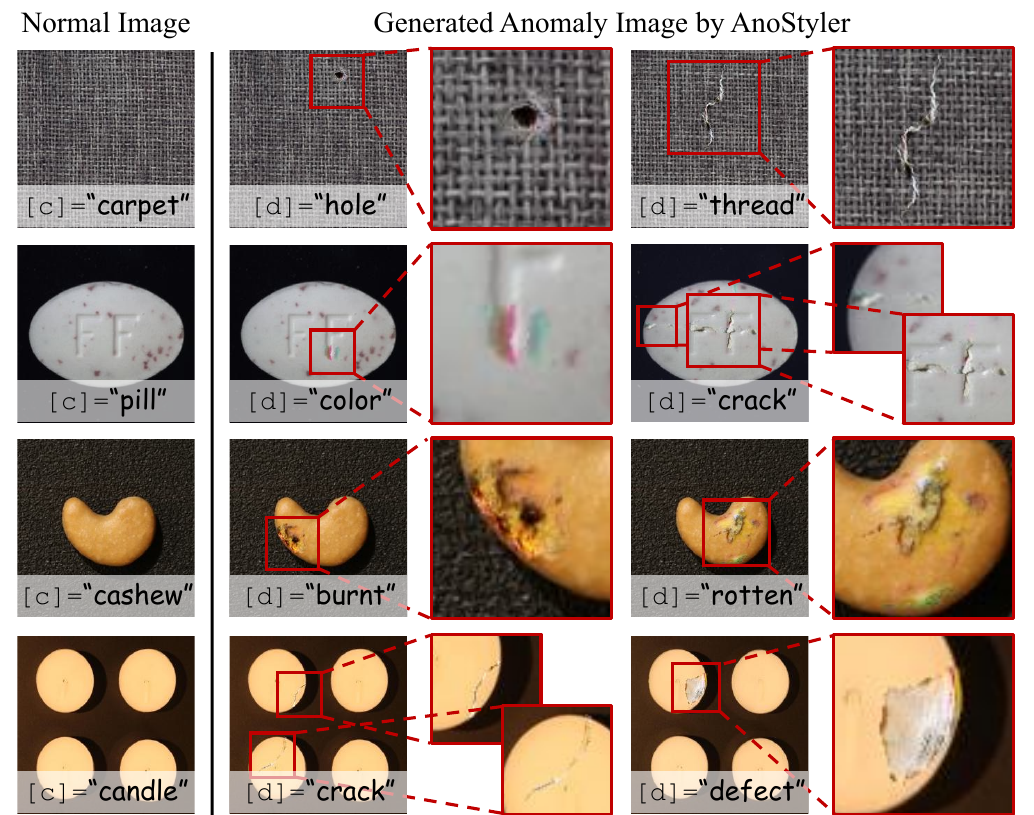}
\caption{Examples of anomaly images generated by AnoStyler. Given a normal image and a category-defect pair $(\texttt{[c]},\texttt{[d]})$, our method generates visually realistic and semantically aligned anomalies.} 
\label{fig:intro_figure}
\end{figure}

Anomaly detection aims to identify patterns or regions in an image that deviate from the learned notion of normality~\cite{li2025survey}. Due to the rarity and diversity of anomalies, unsupervised learning~\cite{defard2021padim,roth2022total,batzner2024efficientad,hyun2024reconpatch,wu2025dfm, fang2025ckaad} on normal images has emerged as the dominant paradigm. Despite their success, these methods lack the capacity to model diverse real-world anomaly distributions, which limits their performance, particularly in complex or unseen domains~\cite{cui2023survey, cao2023anomaldistshift}. 
This underscores the necessity of generating realistic and diverse anomaly images as alternative supervisory signals, leading to active research on \textit{Anomaly Generation} to mitigate the scarcity of real anomaly images.

Existing anomaly generation methods can be broadly categorized into two paradigms: few-shot and zero-shot methods. Few-shot methods typically train generative models to synthesize anomaly images based on a few real anomaly images, assuming access to these real anomalies~\citep{zhang2021defectgan,duan2023defect, hu2024anomalydiffusion, gui2024anogen}. Zero-shot methods, on the other hand, operate under a more challenging yet practical scenario where no anomaly images are available, thus have emerged as promising research directions. 
Earlier zero-shot methods relied on direct manipulation of normal images using handcrafted operations~\cite{lin2021few, li2021cutpaste, zavrtanik2021draem, schluter2022natural}
, whereas more recent methods either employ generative models trained solely on normal images to synthesize anomalies by perturbing the generation process~\citep{zhang2024realnet} or leverage pre-trained generative models guided by text prompts~\citep{sun2025unseen}.

While effective, existing methods still face one or more of the following limitations that restrict their practical applicability. First, the generated anomalies lack visual realism and semantic fidelity, especially for methods that directly manipulate normal images \cite{li2021cutpaste, zavrtanik2021draem, schluter2022natural}. 
Using such low-quality anomalies may result in poor generalization to real-world anomalies.
Second, the generation process necessitates access to large amounts of real normal images \citep{zhang2024realnet}, and further requires even a small number of real anomalies in the case of few-shot methods \citep{duan2023defect, hu2024anomalydiffusion, gui2024anogen}, making it challenging in scenarios where collecting real data is costly or difficult. Third, the generation process depends on memory-intensive heavyweight architectures like diffusion models \citep{duan2023defect, hu2024anomalydiffusion, gui2024anogen, zhang2024realnet, sun2025unseen}, rendering it impractical for real-time or resource-constrained scenarios.

In this work, we propose \textit{AnoStyler}, a lightweight yet effective zero-shot anomaly generation method that simultaneously addresses the aforementioned limitations. AnoStyler frames anomaly generation as a text-driven style transfer task, in which a normal image is transformed into an anomaly image by locally modifying its visual attributes while preserving its overall structural content. Compared to mainstream methods that leverage generative models~\citep{goodfellow2014generative, ho2020ddpm}, style transfer~\cite{gatys2016image} can better preserve the overall content of the original image while injecting localized anomalies, making it inherently more suitable for anomaly generation. 
Nevertheless, it remains underexplored in this context; to the best of our knowledge, we present the first approach that effectively leverages style transfer for this purpose.
AnoStyler integrates shape-guided masking and state-aware prompt generation with tailored losses to achieve semantically aligned anomaly stylization.
As illustrated in Figure~\ref{fig:intro_figure}, AnoStyler generates realistic and diverse anomalies that are semantically consistent with text prompts while preserving the global structure of the input image. Leveraging these components, AnoStyler achieves state-of-the-art performance among zero-shot methods on MVTec-AD and VisA, and its generated anomalies significantly enhance downstream anomaly detection.

Our main contributions are summarized as follows:
\begin{itemize}
    \item We propose a method that generates high-quality and realistic anomalies in various types, with precise semantic alignment to text prompts.
    \item We design a zero-shot anomaly generation framework that requires only a single normal image to synthesize each anomaly, removing the dependency on large collections of normal or anomaly images that hinder scalability.
    \item We introduce a lightweight model architecture that enables computationally and memory-efficient anomaly generation  while preserving competitive performance.
\end{itemize}

\begin{figure*}[!t]
\centering
\includegraphics[width=\linewidth]{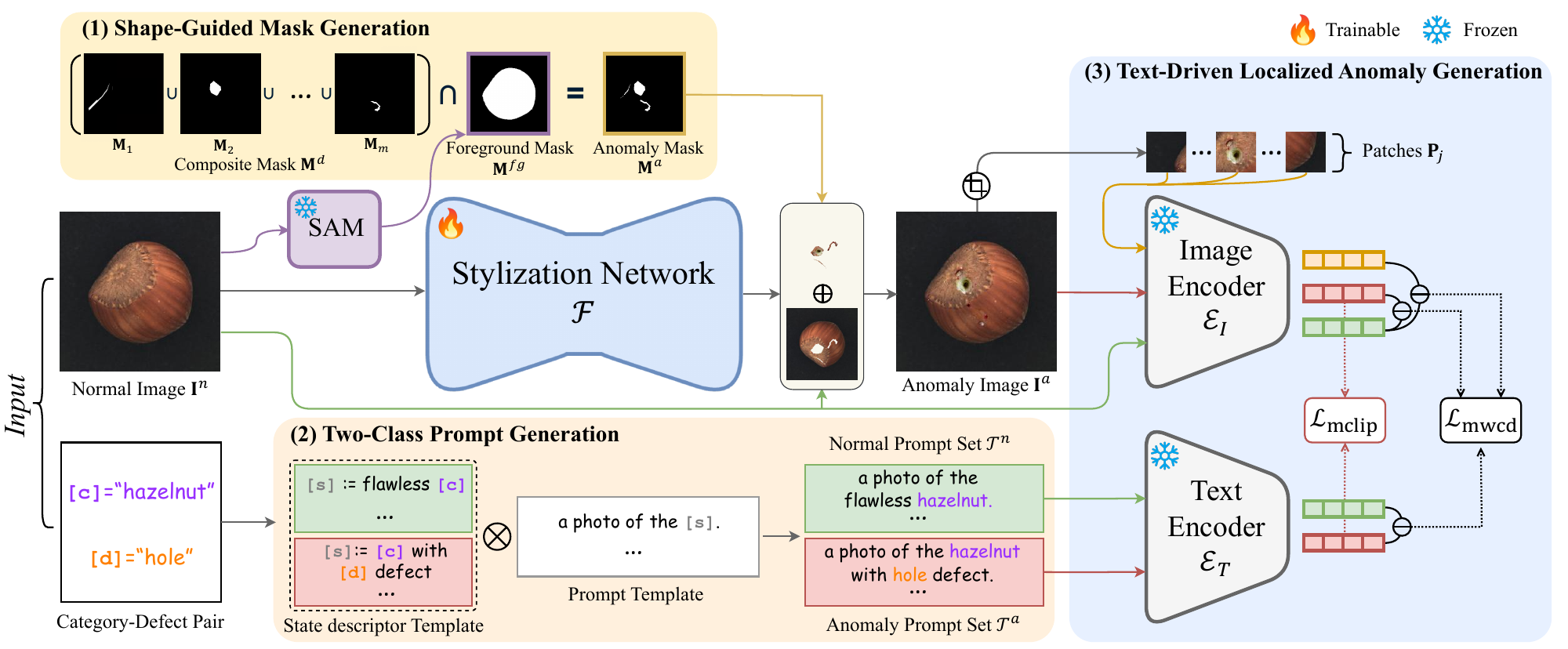}

\caption{Overall framework of AnoStyler. (1) \textbf{Shape-Guided Mask Generation}: A union of primitive masks $\mathbf{M}_1,\ldots,\mathbf{M}_m$ from Meta-Shape Priors is intersected with the foreground mask $\mathbf{M}^{fg}$ to obtain the anomaly mask $\mathbf{M}^a$. 
(2) \textbf{Two-Class Prompt Generation}: Structured text prompt templates are filled with the category-defect pair (\texttt{[c]}, \texttt{[d]}) to form normal and anomaly prompt sets $\mathcal{T}^n$ and $\mathcal{T}^a$. 
(3) \textbf{Text-Driven Localized Anomaly Generation}: Guided by $\mathbf{M}^a$, $\mathcal{T}^n$, and $\mathcal{T}^a$, the stylization network $\mathcal{F}$ is trained to stylize the masked regions of the input image $\mathbf{I}^n$ as anomalies, resulting in the synthetic anomaly image $\textbf{I}^a$.}
\label{fig:framework}
\end{figure*}

\section{Related Work}

\subsection{Anomaly Generation}

Depending on the availability of anomalies in the training data, anomaly generation can be formulated as either a few-shot or zero-shot task.
Few-shot methods leverage a small number of anomaly images to train generative models, 
such as DFMGAN~\citep{duan2023defect} based on generative adversarial networks (GANs)~\citep{goodfellow2014generative}, and AnoDiff~\citep{hu2024anomalydiffusion} and AnoGen~\citep{gui2024anogen} based on diffusion models~\citep{ho2020ddpm}, 
which are trained using a large number of normal images in addition to the few anomaly images.

In contrast, zero-shot methods operate under the assumption that no anomaly images are available. Heuristic methods generate synthetic anomalies by directly manipulating normal images using handcrafted operations. For example, CutPaste~\citep{li2021cutpaste} and NSA~\citep{schluter2022natural} use cut-and-paste operation, and DRAEM~\citep{zavrtanik2021draem} uses external texture injection. Recent studies have proposed methods that leverage generative models to produce more realistic anomalies with enhanced semantic alignment and visual fidelity. RealNet~\citep{zhang2024realnet} utilizes a diffusion model trained on normal images and generates anomalies by perturbing the denoising process. AnomalyAny~\citep{sun2025unseen} guides Stable Diffusion using text prompts to generate anomaly images. While they are effective, they often incur high computational costs and memory usage due to their reliance on diffusion models.

\subsection{Style Transfer}

Style transfer aims to generate a new image that combines the structural content of an input with the texture or style of a reference.
The seminal work on neural style transfer~\cite{gatys2016image} presented an optimization-based method that uses feature statistics from a pre-trained CNN to separately model content and style. Subsequent methods improved efficiency and generality by introducing feature-level transformations, such as AdaIN~\cite{huang2017adain} and WCT~\cite{li2017universal, li2018closed}.

With the advent of CLIP~\cite{radford2021clip}, a new line of research has emerged that replaces reference style images with natural language prompts. StyleCLIP~\cite{patashnik2021styleclip} maps text embeddings to latent directions in StyleGAN~\cite{karras2019stylegan}, enabling text-driven manipulation within the learned image manifold. CLIPstyler~\cite{kwon2022clipstyler} performs text-conditioned style transfer by optimizing patch-wise CLIP loss, achieving localized and semantically meaningful stylization without style images.
Building on CLIPstyler, subsequent research has advanced text-driven and object-centric style editing, exploring approaches for fine-grained and localized stylization. Recent studies~\cite{kamra2023semcs,ganugula2023mosaic,singh2024least,chen2024soulstyler} incorporate mechanisms such as foreground–background separation, segmentation masks, and semantic guidance to selectively apply distinct styles to specific regions or objects.

Following this line of research, we apply text-driven localized style transfer to zero-shot anomaly generation. With CLIP-based losses, AnoStyler efficiently adds text-guided localized anomalies while preserving overall content.

\section{Method}

\subsection{Overview of AnoStyler}
\label{sec:overview}


AnoStyler frames anomaly generation as a text-guided style transfer process that transforms a normal image by injecting localized anomalies. The inputs to the generation process are a real normal image $\textbf{I}^n \in \mathbb{R}^{C \times H \times W}$, where $C$, $H$, and $W$ denote the number of channels, image height, and width, respectively, as well as its category label $\texttt{[c]}$ and expected defect type $\texttt{[d]}$ in text format. Using the image $\textbf{I}^n$ as the content and the category-defect pair $(\texttt{[c]}, \texttt{[d]})$ as a reference style, the generation process outputs a generated anomaly image $\textbf{I}^a \in \mathbb{R}^{C \times H \times W}$.
The generation process, which generates one synthetic anomaly image at a time from one real normal image, consists of three sequential steps, as illustrated in Figure~\ref{fig:framework}.

\subsection{Model Architecture}

The stylization network $\mathcal{F}$ is modeled as a lightweight U-Net~\cite{kwon2022clipstyler} that takes a normal image $\mathbf{I}^n$ as input and generates a stylized output $\mathbf{I}^a$ serving as a synthetic anomaly image.
A pre-trained CLIP model~\cite{radford2021clip} is incorporated to guide anomaly generation by $\mathcal{F}$. The text encoder $\mathcal{E}_T$ is a Transformer encoder that embeds text prompts in $\mathcal{T}^n$ or $\mathcal{T}^a$. The image encoder $\mathcal{E}_I$ is a Vision Transformer that embeds images $\mathbf{I}^n$, $\mathbf{I}^a$, or patches $\mathbf{P}_j$ extracted from $\mathbf{I}^a$. These embeddings are used to compute CLIP-based loss functions that enforce semantic consistency between the generated image and the given text prompt.
During training, only the stylization network $\mathcal{F}$ is trainable, while all parameters of the text encoder $\mathcal{E}_T$ and image encoder $\mathcal{E}_I$ are frozen.

\subsection{Shape-Guided Mask Generation}
\label{sec:maskgeneration}

We argue that a practical anomaly mask generation method should meet two essential criteria. First, it should produce visually plausible masks that generalize across diverse categories and domains without relying on category-specific assumptions. Previous methods often rely on simple geometric shapes like rectangles~\cite{li2021cutpaste,schluter2022natural} or Perlin noise~\cite{zavrtanik2021draem,zhang2024realnet}, which lack the fidelity required to mimic plausible anomalies. Second, it should be lightweight for scalable deployment. Previous methods that rely on dedicated heavyweight model like diffusion models~\cite{duan2023defect,hu2024anomalydiffusion} are not scalable and introduce significant computational overhead. 
To this end, we propose a generalizable category-agnostic and lightweight non-parametric procedure that leverages primitive geometric components.

\paragraph{Meta-Shape Priors.}
For the procedural generation of anomaly masks, we introduce meta-shape priors, which comprise three primitive shapes designed to capture a distinct class of geometric patterns for localized anomaly regions: \textit{Line}, \textit{Dot}, and \textit{Freeform}. These three shape types are carefully chosen to cover a broad spectrum of plausible anomaly geometries observed in real-world data, while agnostic to specific categories or domains. 
Line-shaped masks are created by connecting multiple points along a sampled direction with variable thickness. Sinusoidal perturbations add curvature, effectively capturing straight or wavy line-shaped anomalies.
Dot-shaped masks are generated by sampling points radially around a center and perturbing their radii with noise. They preserve a circular form while producing smooth, spiky, or irregular closed shapes.
Freeform masks are produced by simulating unconstrained random trajectories. Unlike the structured forms above, they generate irregular, topology-free regions resembling diffused or amorphous anomalies.
Detailed procedures for generating masks of these three types are provided in Appendix~\ref{appendix:msp}.

\paragraph{Anomaly Mask Generation.}

To generate an anomaly mask for the input image $\mathbf{I}^n$, we first combine multiple anomaly regions derived from meta-shape priors, then retain the parts that overlap with the foreground of the image. The assumption is that multiple anomaly regions may co-occur within a single image and can appear only in the foreground.

The number of anomaly regions $m$ is sampled from a categorical distribution with exponentially decaying probabilities. The probability mass function is defined as:
\begin{equation}
P(m=i) = \frac{\exp(-\alpha i)}{\sum_{j=1}^{m_\text{max}} \exp(-\alpha j)}, \quad i \in \{1, 2, \dots, m_\text{max}\},
\label{eq:pi}
\end{equation}
where $m_\text{max}$ denotes the maximum number of anomaly regions and $\alpha$ is a decay coefficient that favors fewer anomalies while allowing for more complex compositions to occasionally occur within an image. 

To generate a primitive mask $\mathbf{M}_i$ indicating an anomaly region, we randomly select one of the three meta-shape priors (\textit{i.e.}, Line, Dot, or Freeform) and follow its mask generation procedure. After obtaining $m$ masks, we take their union to allow multiple distinct regions to be represented within a single mask. The composite mask $\mathbf{M}^d$ is given by:
\begin{equation}
\mathbf{M}^d = \bigcup_{i=1}^{m} \mathbf{M}_i.
\label{eq:maskunion}
\end{equation}

The final anomaly mask $\mathbf{M}^a$ is obtained by intersecting $\mathbf{M}^d$ with the foreground mask $\mathbf{M}^{fg}$ indicating the foreground region of an image. 
If the category label \texttt{[c]} for
an image given is object-centric, where anomalies typically occur on object surfaces, we use the Segment Anything Model (SAM)~\cite{kirillov2023segment} 
with a ViT-B backbone to generate the foreground mask. Specifically, we designate four corner points as positive prompts to guide SAM in extracting the background region, and then apply a negation operation to obtain the foreground mask $\mathbf{M}^{fg}$.
If the category label \texttt{[c]} is texture-centric, such as fabrics or surfaces without distinct object boundaries, we treat the entire image as foreground by setting $\mathbf{M}^{fg} = \mathbf{1}^{H \times W}$. The final anomaly mask $\mathbf{M}^a$ is then given by:
\begin{equation}
    \mathbf{M}^a = \mathbf{M}^d \cap \mathbf{M}^{fg}.
\end{equation}

\subsection{Two-Class Prompt Generation}
\label{sec:dualprompt}

We propose a generalizable template-based text prompt generation procedure by adapting three key techniques: two-class design, prompt templates, and prompt averaging. Two-class design~\citep{jeong2023winclip} constructs separate text prompts for the normal and anomaly states. Explicitly modeling each state provides clearer guidance on how an anomaly image should differ from a normal one. The use of structured prompt templates facilitates intuitive and scalable prompt generation by requiring only the specification of the category label and defect type, without requiring domain-specific knowledge or real anomaly images.  
Prompt averaging~\citep{radford2021clip} generates multiple semantically equivalent prompts for the given context and averages their embeddings to guide anomaly generation. This reduces variability and potential bias introduced by any single prompt, resulting in more stable and consistent text conditioning.

Given the category-defect pair (\texttt{[c]}, \texttt{[d]}) for the input image $\mathbf{I}^n$, we first generate two-class state descriptors \texttt{[s]} for the normal and anomaly states using predefined state descriptor templates~\cite{jeong2023winclip}. For example, the normal state is described by phrases such as \texttt{"flawless [c]"}, while the anomaly state is described by phrases such as \texttt{"[c] with [d] defect"}. Next, each state descriptor \texttt{[s]} is inserted into predefined prompt templates to form full text prompts such as \texttt{"a photo of the [s]"}~\cite{kwon2022clipstyler}. If the category label \texttt{[c]} or the defect type \texttt{[d]} is not provided, the default token \texttt{"sample"} or \texttt{"defect"} can be used, respectively. This yields a normal prompt set $\mathcal{T}^n$ and an anomaly prompt set $\mathcal{T}^a$. The complete list of templates used is provided in Appendix~\ref{appendix:template}. By prompt averaging, the embeddings of $\mathcal{T}^n$ and $\mathcal{T}^a$ are obtained using the clip encoder $\mathcal{E}_T$ as:
\begin{equation}
\mathbf{h}_T^n = \frac{1}{|\mathcal{T}^n|} \sum_{T_i \in \mathcal{T}^n} \mathcal{E_T}(T_i);\quad\mathbf{h}_T^a = \frac{1}{|\mathcal{T}^a|} \sum_{T_i \in \mathcal{T}^a} \mathcal{E_T}(T_i).
\end{equation}

\subsection{Text-Driven Localized Anomaly Generation}
\label{sec:methodoverview}

The anomaly mask $\mathbf{M}^a$ and the prompt sets $\mathcal{T}^n$ and $\mathcal{T}^a$ are used to guide the stylization network $\mathcal{F}$ to transform the input image $\mathbf{I}^n$ by incorporating localized anomalies. We introduce two loss terms specifically designed to promote the spatial localization and semantic alignment of synthetic anomalies: the \textit{Mask-Weighted Co-Directional Loss} $\mathcal{L}_{\text{mwcd}}$ and the \textit{Masked CLIP Loss} $\mathcal{L}_{\text{mclip}}$. Following CLIPstyler, we additionally adopt the \textit{Content Loss} $\mathcal{L}_{c}$~\cite{gatys2016image} and the \textit{Total Variation Loss} $\mathcal{L}_{\text{tv}}$~\cite{rudin1992nonlinear} to ensure the semantic fidelity and spatial smoothness of the generated image. The training objective for $\mathcal{F}$ is given by:
\begin{equation}
\mathcal{L} = \mathcal{L}_{\text{mwcd}}
    + \lambda_{\text{mclip}} \cdot \mathcal{L}_{\text{mclip}}
    + \lambda_{\text{c}} \cdot \mathcal{L}_{\text{c}}
    + \lambda_{\text{tv}} \cdot \mathcal{L}_{\text{tv}},
\end{equation}
where $\lambda_{\text{mclip}}$, $\lambda_{\text{c}}$, and $\lambda_{\text{tv}}$ are weights that control the strengths of each loss term.

After the stylization network $\mathcal{F}$ is trained, the stylized image $\mathbf{I}^a$, which serves as a generated anomaly image, is generated by compositing the network output $\mathcal{F}(\mathbf{I}^n)$ with the original input image $\mathbf{I}^n$ using the anomaly mask $\mathbf{M}^a$ as a binary weight matrix: 
\begin{equation}
\mathbf{I}^a = \mathcal{F}(\mathbf{I}^n) \odot\mathbf{M}^a + \mathbf{I}^n \odot (1 - \mathbf{M}^a),
\label{eq:stylemask}
\end{equation}
where $\odot$ is the element-wise product operator. This ensures that style transfer is applied exclusively within the masked regions while preserving the surrounding background.

\paragraph{Mask-Weighted Co-Directional Loss.}  

The concept of directional supervision in CLIP embedding space, originally introduced in StyleGAN-NADA~\cite{gal2021stylegannada} and further explored in CLIPstyler~\cite{kwon2022clipstyler}, is the basis of the loss 
$\mathcal{L}_{\text{mwcd}}$. The loss is computed as a weighted sum of a global directional alignment term $\mathcal{L}_{\text{gdir}}$ and a patch-wise directional alignment term $\mathcal{L}_{\text{pdir}}$, defined as:
\begin{equation}
\mathcal{L}_{\text{mwcd}} = \lambda_{\text{gdir}} \cdot \mathcal{L}_{\text{gdir}} + \lambda_{\text{pdir}} \cdot \mathcal{L}_{\text{pdir}},
\end{equation}
where $\lambda_{\text{gdir}}$ and $\mathcal{L}_{\text{pdir}}$ are weights assigned to each term. We modify the original loss to place greater emphasis on anomaly regions indicated by the anomaly mask $\mathbf{M}^a$. The modified loss terms incorporating the mask $\mathbf{M}^a$ and the mask-guided anomaly image $\mathbf{I}^a$ are detailed as follows.

The first term $\mathcal{L}_{\text{gdir}}$ measures global directional alignment between images and text prompts by comparing their respective shifts from the normal to the anomaly state in the CLIP embedding space. The directional shifts of the image and prompt embeddings are computed as:
\begin{align}
\Delta \mathbf{h}_I &= \mathbf{h}_I^a-\mathbf{h}_I^n = \mathcal{E}_{I}(\mathbf{I}^a) - \mathcal{E}_{I}(\mathbf{I}^n);\\
\Delta \mathbf{h}_T &= \mathbf{h}_T^a-\mathbf{h}_T^n.
\end{align}
The term $\mathcal{L}_{\text{gdir}}$ is then defined as the cosine distance between $\Delta \mathbf{h}_I$ and $\Delta \mathbf{h}_T$:
\begin{equation}
\mathcal{L}_{\text{gdir}} = 1 - \frac{\Delta \mathbf{h}_I \cdot \Delta \mathbf{h}_T}{\|\Delta \mathbf{h}_I\| \cdot \|\Delta \mathbf{h}_T\|}.
\end{equation}
Minimizing this term encourages the global semantic shift from the normal image $\mathbf{I}^n$ to the generated image $\mathbf{I}^a$ to align with the intended transformation described by the two-class prompt sets $\mathcal{T}^n$ and $\mathcal{T}^a$.

The second term $\mathcal{L}_{\text{pdir}}$ is a patch-wise extension of $\mathcal{L}_{\text{gdir}}$ designed to further refine alignment at a finer scale. It measures patch-wise directional alignment by operating on image patches extracted from $\mathbf{I}^a$, thereby providing localized semantic guidance in contrast to global alignment over the whole image. Here, we extract $l$ patches $\{\mathbf{P}_j\}_{j=1}^l$ from $\mathbf{I}^a$ via random cropping. For each patch $\mathbf{P}_j$, we measure the local directional shift from the input image $\mathbf{I}^n$ as:
\begin{equation}
\Delta \mathbf{h}_{P_j} = \mathbf{h}^a_{P_j} - \mathbf{h}_{I}^n =\mathcal{E}_{I}(\tau(\mathbf{P}_j)) - \mathcal{E}_{I}(\mathbf{I}^n),
\end{equation}
where $\tau$ denotes a random perspective transformation applied to induce patch-wise geometric variation. The term $\mathcal{L}_{\text{pdir}}$ is then defined as:
\begin{equation}
\mathcal{L}_{\text{pdir}} = \frac{1}{{\sum_{j=1}^{l} r_j}} \text{ } {\sum_{j=1}^{l} r_j  \left( 1 - \frac{\Delta \mathbf{h}_{P_j} \cdot \Delta \mathbf{h}_T}{\|\Delta \mathbf{h}_{P_j}\| \cdot \|\Delta \mathbf{h}_T\|} \right)},
\end{equation}
where $r_j \in [0, 1]$ denotes the ratio of pixels in $\mathbf{P}_j$ that are covered by the anomaly mask $\mathbf{M}^a$. With this soft weighting, each patch $\mathbf{P}_j$ contributes proportionally to its overlap with $\mathbf{M}^a$, thereby placing greater emphasis on anomaly regions.

\begin{table}[!t]
\centering
\tabcolsep=6pt
\renewcommand{\arraystretch}{1.0}

\resizebox{\linewidth}{!}{\begin{tabular}{l|cc|cc}
\toprule
   &
  \multicolumn{2}{c|}{MVTec-AD} &
  \multicolumn{2}{c}{VisA} \\ \cmidrule(l){2-5} 
 {Method} & IS  & IC-L  & IS  & IC-L  \\
\midrule
\multicolumn{5}{c}{\textit{Few-Shot Anomaly Generation}} \\[-2pt]
\midrule
DFMGAN$^{\star}$~\cite{duan2023defect} &
1.72 & 0.20 &
1.48 & 0.28 \\
AnoDiff$^{\star}$~\cite{hu2024anomalydiffusion} &
1.80 & \underline{0.32} & 1.50 & \underline{0.29} \\
AnoGen$^{\dagger}$~\cite{gui2024anogen} &
1.77 & 0.27 & 1.40 & 0.22 \\
\midrule
\multicolumn{5}{c}{\textit{Zero-Shot Anomaly Generation}} \\[-2pt]
\midrule
CutPaste$^{\dagger}$~\cite{li2021cutpaste} &
1.76 & 0.22 & 1.52 & 0.26 \\
DRAEM$^{\dagger}$~(Zavrtanik et al. 2021) &
1.76 & 0.25 & 1.50 & 0.25 \\
NSA$^{\circ}$~\cite{schluter2022natural} &
1.44 & 0.26 & 1.42 & 0.19 \\
RealNet$^{\circ}$~(Zhang et al. 2024) &
1.64 & 0.22 & \underline{1.53} & 
\underline{0.29} \\
AnomalyAny$^{\circ}$~\cite{sun2025unseen} &
\underline{2.02} & \textbf{0.33} & 1.41 & 0.19 \\
\midrule
\textbf{{{AnoStyler}} (ours)}&
\textbf{2.04} & \underline{0.32} & \textbf{1.55} & \textbf{0.32} \\
\bottomrule
\end{tabular}}
\caption{Comparison of anomaly generation on MVTec-AD and VisA. For each metric, the best and second-best scores are shown in \textbf{bold} and \underline{underlined}.
${\dagger}$: Re-implemented on both datasets; ${\star}$ and ${\circ}$: Results on MVTec-AD taken from~\cite{hu2024anomalydiffusion} and~\cite{sun2025unseen}, respectively, with results on VisA re-implemented for a fair comparison.}
\label{tab:generation}
\end{table}

\paragraph{Masked CLIP Loss.}  
The loss $\mathcal{L}_{\text{mclip}}$ is designed to further enforce semantic alignment between the localized anomalies in the generated anomaly image $\mathbf{I}^a$ and the target semantics described by the anomaly prompt set $\mathcal{T}^a$. Specifically, the loss is defined as the cosine distance between the masked region of the generated image $\mathbf{I}^a$ and the anomaly prompt set $\mathcal{T}^a$ in the CLIP embedding space:
\begin{equation}
\mathcal{L}_{\text{mclip}} = 1 - \frac{\mathcal{E}_{I}(\mathbf{I}^a \odot \mathbf{M}^a) \cdot \mathbf{h}_T^a}{\|\mathcal{E}_{I}(\mathbf{I}^a \odot \mathbf{M}^a)\| \cdot \|\mathbf{h}_T^a\|}.
\end{equation}
Restricting this loss to the regions specified by the anomaly mask $\mathbf{M}^a$ ensures that semantic alignment is focused only on these anomaly regions, without affecting the background of the input image.

\begin{figure*}[!t]
\centering
\includegraphics[width=\linewidth]{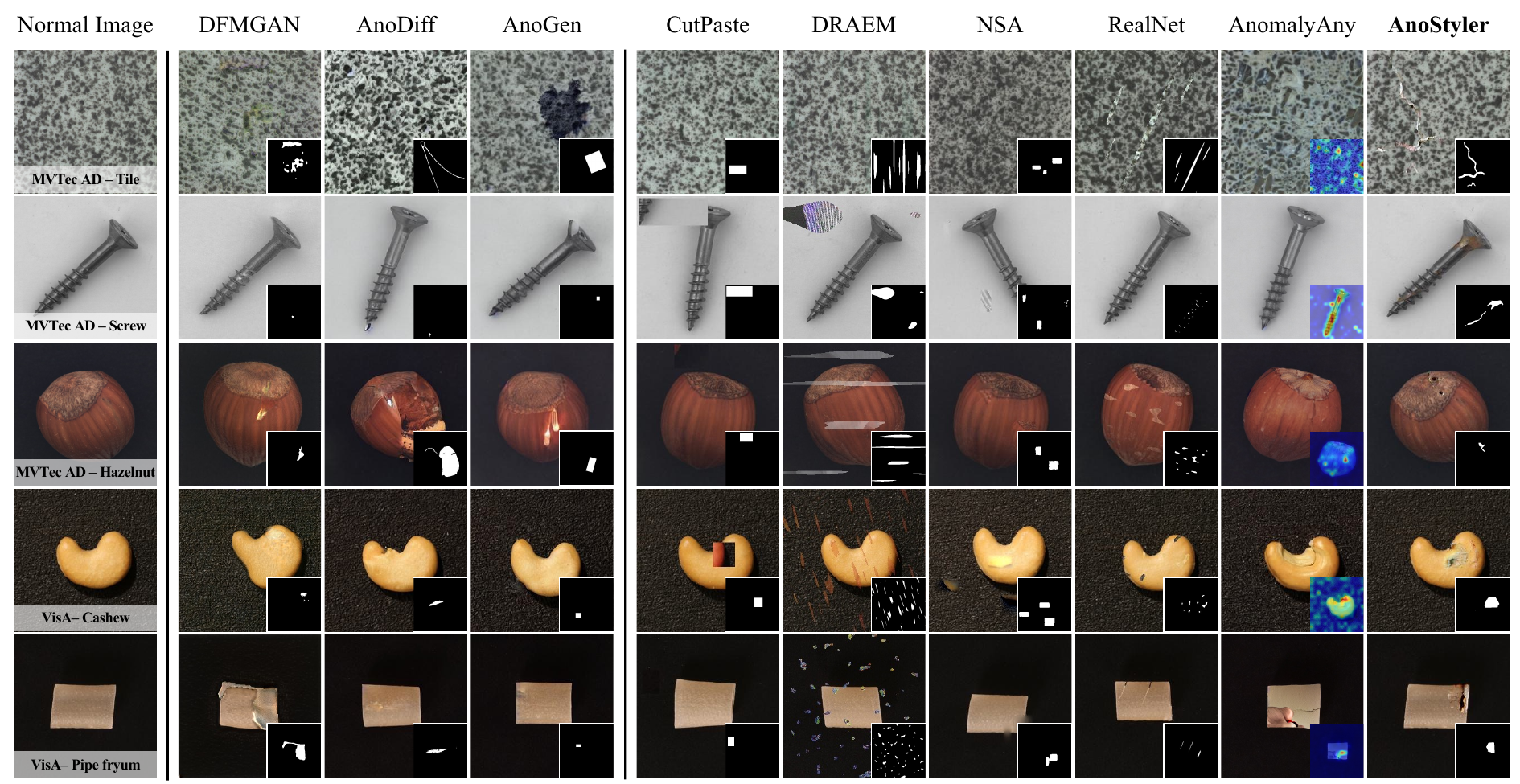}
\caption{Comparison of generated anomaly images and their corresponding anomaly masks on MVTec-AD and VisA. AnoStyler generates visually realistic anomalies that align well with the masks.}
\label{fig:gen_qual}
\end{figure*}

\section{Experiments}
\subsection{Experiment Settings}

\paragraph{Datasets.}

We conducted experiments on two representative benchmark datasets for industrial visual anomaly detection: MVTec-AD~\cite{bergmann2019mvtec} and VisA~\cite{zou2022spot}. MVTec-AD consists of 5,354 high-resolution images across 10 object and 5 texture categories, each associated with 1 to 7 defect types, which is curated to support the detection of subtle local anomalies in controlled settings.
VisA consists of 10,821 images spanning 12 object categories, and it captures more complex scenes with multi-object arrangements and diverse structural variations.
For both datasets, each image is paired with a ground-truth pixel-wise anomaly mask. In the experiments, all images and masks were resized to 512$\times$512. Detailed information about the benchmark datasets is provided in Appendix~\ref{appendix:experiment_details}.

\paragraph{Implementation Details.}

The stylization network $\mathcal{F}$ had a U-Net architecture, consisting of three downsampling blocks and three upsampling blocks, as used in \citet{kwon2022clipstyler}. For the image encoder $\mathcal{E}_I$ and text encoder $\mathcal{E}_T$, we used the pre-trained CLIP ViT-B/32 model.
Detailed hyperparameter configurations of AnoStyler are provided in Appendix~\ref{appendix:hyperparam}. All experiments were conducted on a single NVIDIA RTX 2080Ti GPU with 11 GB of memory. All experiments were repeated five times with different random seeds, and the average performance is reported.

\paragraph{Baselines.}
Our method AnoStyler was compared with a diverse set of zero-shot anomaly generation methods, including CutPaste~\cite{li2021cutpaste}, 
DRAEM~\cite{zavrtanik2021draem}, NSA~\cite{schluter2022natural}, RealNet~\cite{zhang2024realnet}, and AnomalyAny~\cite{sun2025unseen}. For broader comparison, we also considered few-shot methods, including DFMGAN~\cite{duan2023defect}, 
AnoDiff~\cite{hu2024anomalydiffusion},
AnoGen~\cite{gui2024anogen}. It should be noted that few-shot methods have an inherent advantage in terms of performance, as they benefit from access to a small number of real anomaly images.

\begin{table*}[t]
\centering
\tabcolsep=5pt
\renewcommand{\arraystretch}{1.0}

\resizebox{\linewidth}{!}{\begin{tabular}{l|ccccccc|ccccccc}
\toprule
 &
\multicolumn{7}{c|}{MVTec-AD} & 
\multicolumn{7}{c}{VisA} \\
\cmidrule(lr){2-15}

Method& I-AUC & I-AP & I-F1 & P-AUC & P-AP & P-F1 & PRO &
I-AUC & I-AP & I-F1 & P-AUC & P-AP & P-F1 & PRO \\
\midrule
\multicolumn{15}{c}{\textit{Few-Shot Anomaly Generation}} \\[-2pt]
\midrule
DFMGAN$^{\star}$~\cite{duan2023defect} &
87.2 & 94.8 & 94.7 & 90.0 & 62.7 & 62.1 & 76.3 &
83.7 & 85.7 & 80.3 & 90.6 & 31.0 & 34.6 & 74.9 \\
AnoDiff$^{\star}$~\cite{hu2024anomalydiffusion} &
99.2 & 99.7 & 98.7 & 99.1 & 81.4 & 76.3 & 94.0 &
86.9 & 89.1 & 85.4 & 93.2 & 33.0 & 36.8 & 79.0 \\
AnoGen$^{\dagger}$~\cite{gui2024anogen} &
98.7& 99.6& 97.7& 96.9& 73.6& 66.7& 90.7&
90.4 & 92.4 & 87.2 & 89.1 & 31.6 & 33.4 & 73.8 \\
\midrule
\multicolumn{15}{c}{\textit{Zero-Shot Anomaly Generation}} \\[-2pt]
\midrule
CutPaste$^{\dagger}$~\cite{li2021cutpaste} &
89.8 & 92.1 & 89.8 & 88.2 & 51.9 & 50.7 & 76.4 &
86.3 & 86.9 & 87.1 & 88.4 & \underline{32.2} & \underline{39.6} & 77.7 \\
DRAEM$^{\star}$~(Zavrtanik et al. 2021) &
94.6 & \underline{97.0} & 94.4 & 92.2 & 54.1 & 53.1 & 83.1 &
91.8&92.9&88.6&91.4&29.5&37.2&81.9\\
NSA$^{\dagger}$~\cite{schluter2022natural} &
93.0&95.6&91.6&92.0&52.6&52.5&82.2&
87.3&89.8&84.2&\underline{92.6}&26.1&34.2&74.2\\
RealNet$^{\dagger}$~(Zhang et al. 2024) &
\underline{95.2} & \underline{97.0} & 95.3 & \underline{94.0} & 57.7 & 56.6 & \underline{85.2} &
\underline{92.6} & \underline{93.8} & \underline{89.2} & 92.2 & \textbf{33.3} & \textbf{41.0} & 83.0 \\
AnomalyAny$^{\dagger}$~\cite{sun2025unseen} &
\underline{95.2} & 96.9 & \underline{96.3} & 89.0 & \underline{62.7} & \underline{59.9} & 84.7 &
88.9 & 86.2 & 85.9 & 90.4 & 31.2 & 33.0 & \textbf{84.6} \\
\midrule
\textbf{{AnoStyler} (ours)} &
\textbf{98.0} & \textbf{99.0} & \textbf{97.0} & \textbf{94.4} & \textbf{62.9} & \textbf{60.7} & \textbf{88.3} &
\textbf{93.9} & \textbf{95.3} & \textbf{90.1} & \textbf{93.8} & {31.4} & {36.4} & \underline{84.3} \\
\bottomrule
\end{tabular}}
\caption{Comparison of anomaly detection on MVTec-AD and VisA. All scores are shown in percentages. For each metric, the best and second-best scores among zero-shot methods are indicated in \textbf{bold} and \underline{underlined}. ${\dagger}$: Re-implemented on both datasets; ${\star}$: Results on MVTec-AD taken from~\cite{hu2024anomalydiffusion}, with results on VisA re-implemented for a fair comparison.}
\label{tab:detection}
\end{table*}

\subsection{Anomaly Generation Results}
\label{sec:generation_results} 

We first evaluated the effectiveness of AnoStyler and the baseline methods in anomaly generation. 
Following the evaluation protocol of~\cite{hu2024anomalydiffusion}, we generated 1,000 anomaly images per category for each method. 
The Inception Score (IS)~\cite{salimans2016improved} and the Intra-Cluster pairwise LPIPS distance (IC-L)~\cite{ojha2021fewshot} were used to quantitatively evaluate generation quality and diversity.

Table~\ref{tab:generation} presents the quantitative performance comparison for anomaly generation, with results averaged across all categories within each benchmark dataset. The full category-wise results of AnoStyler are provided in Appendix~\ref{appendix:E}. AnoStyler achieved the highest IS and the second-highest IC-L on MVTec-AD, and obtained the best scores for both metrics on VisA, indicating superior visual quality and diversity in its images.
The qualitative results are presented in Figure~\ref{fig:gen_qual}.
The images generated by AnoStyler demonstrate competitive visual realism compared to few-shot methods that have access to real anomaly images, including DFMGAN, AnoDiff, and AnoGen. In contrast, heuristic zero-shot methods, including CutPaste, DRAEM, and NSA, tend to produce unrealistic anomalies. Among generative zero-shot methods, RealNet also suffers from a lack of realism, while AnomalyAny generates more realistic anomalies but often introduces artifacts and overly smoothed details in certain categories. Additional qualitative examples for all categories generated by AnoStyler are provided in Appendix~\ref{appendix:F}.

\subsection{Anomaly Detection Results}
\label{sec:detection_results} 

We evaluated the methods based on their impact on the performance of the downstream anomaly detection task
The evaluation protocol followed those used in \citet{zavrtanik2021draem} and \citet{hu2024anomalydiffusion}. For each category, we generated 500 anomaly images and used them with real normal images to train a U-Net model, which takes an image as input and predicts its anomaly mask. 
Anomaly detection performance was assessed on the test sets from each benchmark dataset. 
We used Area Under ROC Curve (AUROC), Average Precision (AP), and F1-score at both the image level (I-AUC, I-AP, I-F1) and pixel level (P-AUC, P-AP, P-F1)
. Additionally, we included Per-Region Overlap (PRO) to evaluate region-level localization performance.

Table~\ref{tab:detection} presents the results averaged across categories within each benchmark dataset. The full category-wise results of AnoStyler are provided in Appendix~\ref{appendix:E}.
AnoStyler achieved state-of-the-art anomaly detection 
performance on both MVTec-AD and VisA under zero-shot settings. Remarkably, despite having no access to real anomaly images during the generation of synthetic anomalies, AnoStyler yielded performance comparable with or even superior to few-shot baselines. This highlights the effectiveness of AnoStyler for downstream anomaly detection. Additional statistical significance analysis is provided in Appendix~\ref{appendix:G}.

\begin{table}[!t]
\centering
\tabcolsep=6pt
\renewcommand{\arraystretch}{1.0}
\resizebox{\linewidth}{!}{  
\begin{tabular}{c|ccc|cc|cc}
\toprule
 \#
  & \shortstack{$\mathcal{L}_\text{gdir}$} & \shortstack{$\mathcal{L}_\text{pdir}$} & \shortstack{$\mathcal{L}_\text{mclip}$} & IS  & IC-L  & I-AUC & P-AUC  \\
\midrule
(a) &  &  &  & 1.70 & 0.25 & 88.2 & 85.7 \\
(b) & \checkmark &  &  & 1.86 & 0.29 & 95.2 & 92.5 \\
(c) & \checkmark & \checkmark &  & 1.96 & 0.30 & 96.7 & 93.2 \\
(d) & \checkmark & \checkmark & \checkmark & \textbf{2.04} & \textbf{0.32} & \textbf{98.0} & \textbf{94.4} \\
\bottomrule
\end{tabular}}
\caption{Ablation study on the effect of the proposed loss components in AnoStyler on the MVTec-AD dataset. The results on the VisA dataset are provided in Appendix~\ref{appendix:E}.}
\label{tab:loss_ablation}
\end{table}

\begin{figure}[!t]
\centering
\includegraphics[width=\linewidth]{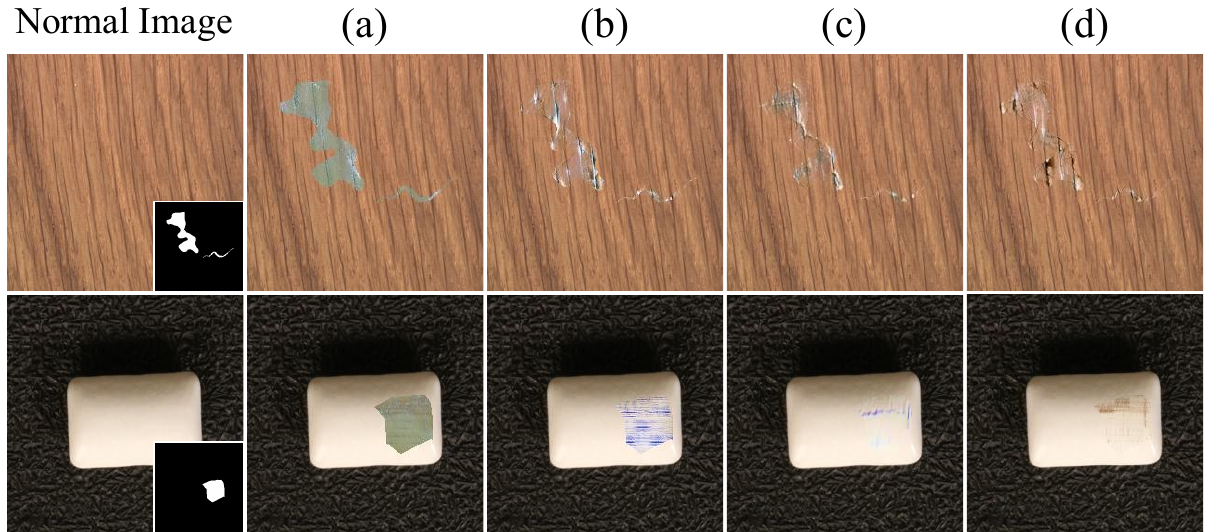}
\caption{Qualitative results corresponding to the loss configurations presented in Table~\ref{tab:loss_ablation}.
The first and second rows use category labels $\texttt{[c]} = \texttt{"wood"}$ and $\texttt{"chewinggum"}$, respectively, with defect type $\texttt{[d]} = \texttt{"scratch"}$.}
\label{fig:loss_ablation}
\end{figure}

\subsection{Ablation Study}
\label{sec:ablation_study}

We conducted an ablation study to analyze the contributions of the proposed loss components: modifying $\mathcal{L}_{\text{gdir}}$, modifying $\mathcal{L}_{\text{pdir}}$, and adding $\mathcal{L}_{\text{mclip}}$, relative to the baseline training objective~\cite{kwon2022clipstyler}. Table~\ref{tab:loss_ablation} presents the quantitative results. Directly applying the baseline objective, denoted as (a), resulted in degraded performance. The proposed modifications
to $\mathcal{L}_{\text{gdir}}$ and $\mathcal{L}_{\text{pdir}}$, as in (b) and (c), led to performance improvements, highlighting their respective contributions. The best performance was achieved with (d), which corresponds to the full AnoStyler training objective incorporating all three proposed components, demonstrating their combined effectiveness. As shown in the qualitative examples in Figure~\ref{fig:loss_ablation}, while (a) produces unrealistic anomalies, (b), (c), and (d) show gradual improvements in both visual realism and semantic alignment with the text prompt.

\subsection{Computational Efficiency}

AnoStyler comprises 263M parameters in total, including 91M for SAM, 0.61M for the stylization network $\mathcal{F}$, 151M for the CLIP encoders $\mathcal{E}_T$ and $\mathcal{E}_I$, and 20M for the feature extractor computing $\mathcal{L}_\text{c}$, whereas diffusion-based baselines such as AnoDiff, AnoGen, RealNet, and AnomalyAny each contain over 1B parameters, making AnoStyler much more compact.
Among these methods, AnomalyAny is the only one that generates anomalies without large amounts of real images, serving as the most comparable baseline. Generating one anomaly image requires 9.5 TFLOPs for AnoStyler versus 22.8 TFLOPs for AnomalyAny, demonstrating its substantially lower computational cost.

\section{Conclusion}

In this paper, we proposed \textit{AnoStyler}, a novel zero-shot anomaly generation method that generates synthetic anomaly images through text-guided style transfer from a single normal image. It produces high-quality, diverse anomalies with strong visual realism and semantic alignment to text prompts, without requiring access to large amounts of real images. In addition, its lightweight architecture enables computationally and memory-efficient anomaly generation. In experimental results on the MVTec-AD and VisA benchmarks, AnoStyler not only outperformed existing zero-shot methods but also achieved performance comparable to few-shot methods in both anomaly generation and downstream anomaly detection. These findings suggest that AnoStyler offers a text-driven alternative to recent mainstream methods that demand substantial computational resources, making it well suited for real-world deployment.

\section*{Acknowledgements}
This work was supported by the National Research Foundation of Korea (NRF) grant (No. RS-2023-00207903) and the Institute of Information \& Communications Technology Planning \& Evaluation (IITP) grant (No. RS-2025-02214591, Development of an Innovative AI Agent for Worker-Friendly Autonomous Manufacturing), funded by the Korea government (MSIT; Ministry of Science and ICT).

\bibliography{aaai2026}

\clearpage

\input{supp}
\end{document}

%% file: supp.tex
\appendix

\onecolumn
\section{Details of Benchmark Datasets}
\label{appendix:experiment_details}

Tables~\ref{tab:defect_types_mvtec} and \ref{tab:defect_types_visa} present the complete list of categories, image counts, and defect types used in MVTec-AD and VisA, respectively. In MVTec-AD, each anomalous image is annotated with a single, well-defined defect type. These predefined defect types were directly used to construct text prompts for guiding AnoStyler’s anomaly generation. In VisA, annotations include multiple, free-form textual descriptions per anomaly image. To maintain consistency, we uniformly assign the label ``defect'' as the defect type for all anomaly images across all categories.

\begin{table*}[!ht]
\centering
\renewcommand{\arraystretch}{1.2}
\small
\setlength{\tabcolsep}{6pt}
\begin{tabularx}{\linewidth}{c|l|r|r|r|X}
\toprule
\textbf{Type} & \textbf{Category} \texttt{[c]} & \makecell{\textbf{\# Train} \\ \textbf{(Normal)}} & \makecell{\textbf{\# Test} \\ \textbf{(Normal)}} & \makecell{\textbf{\# Test} \\ \textbf{(Anomaly)}} & \textbf{Defect Types} \texttt{[d]} \\
\midrule
\multirow{12}{*}{Object}
& bottle        & 209 & 20 & 63 & broken\_large, broken\_small, contamination \\
& cable         & 224 & 58 & 92 & bent\_wire, cable\_swap, combined, cut\_inner\_insulation, cut\_outer\_insulation, missing\_cable, missing\_wire, poke\_insulation \\
& capsule       & 219 & 23 & 109 & crack, faulty\_imprint, poke, scratch, squeeze \\
& hazelnut      & 391 & 40 & 70 & crack, cut, hole, print \\
& metal\_nut     & 220 & 22 & 93 & bent, color, flip, scratch \\
& pill          & 267 & 26 & 141 & color, combined, contamination, crack, faulty\_imprint, pill\_type, scratch \\
& screw         & 320 & 41 & 119 & manipulated\_front, scratch\_head, scratch\_neck, thread\_side, thread\_top \\
& toothbrush    & 60  & 12 & 30  & defective \\
& transistor    & 213 & 60 & 40  & bent\_lead, cut\_lead, damaged\_case, misplaced \\
& zipper        & 240 & 32 & 119 & broken\_teeth, combined, fabric\_border, fabric\_interior, rough, split\_teeth, squeezed\_teeth \\
\midrule
\multirow{5}{*}{Texture}
& carpet        & 280 & 28 & 89 & color, cut, hole, metal\_contamination, thread \\
& grid          & 264 & 21 & 57 & bent, broken, glue, metal\_contamination, thread \\
& leather       & 245 & 32 & 92 & color, cut, fold, glue, poke \\
& tile          & 230 & 33 & 84 & crack, glue\_strip, gray\_stroke, oil, rough \\
& wood          & 247 & 19 & 60 & color, combined, hole, liquid, scratch \\
\bottomrule
\end{tabularx}
\caption{Categories, image counts, and defect types in MVTec-AD.}
\label{tab:defect_types_mvtec}
\end{table*}

\begin{table*}[!ht]
\centering
\renewcommand{\arraystretch}{1.2}
\small
\setlength{\tabcolsep}{6pt}
\begin{tabularx}{\linewidth}{c|l|r|r|r|X}
\toprule
\textbf{Type} & \textbf{Category} \texttt{[c]} & \makecell{\textbf{\# Train} \\ \textbf{(Normal)}} & \makecell{\textbf{\# Test} \\ \textbf{(Normal)}} & \makecell{\textbf{\# Test} \\ \textbf{(Anomaly)}} & \textbf{Defect Types} \texttt{[d]} \\
\midrule
\multirow{12}{*}{Object}
& candle       & 900 & 100 & 100 & defect \\
& capsules     & 542 &  60 & 100 & defect \\
& cashew       & 450 &  50 & 100 & defect \\
& chewinggum   & 453 &  50 & 100 & defect \\
& fryum        & 450 &  50 & 100 & defect \\
& macaroni1    & 900 &  100 & 100 & defect \\
& macaroni2    & 900 &  100 & 100 & defect \\
& pcb1         & 904 &  100 & 100 & defect \\
& pcb2         & 901 &  100 & 100 & defect \\
& pcb3         & 905 &  101 & 100 & defect \\
& pcb4         & 904 &  101 & 100 & defect \\
& pipe\_fryum  & 450 &  50 & 100 & defect \\
\bottomrule
\end{tabularx}
\caption{Categories, image counts, and defect types in VisA.}
\label{tab:defect_types_visa}
\end{table*}

\twocolumn
\section{Meta-Shape Priors}
\label{appendix:msp}

\begin{figure}[!ht]
\centering
\includegraphics[width=\linewidth]{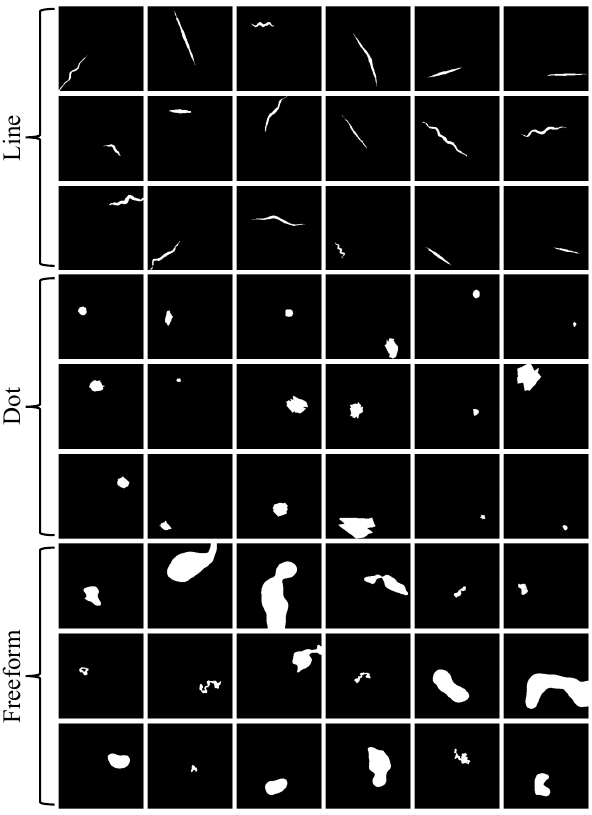}
\caption{Example masks generated by the proposed Meta-Shape Priors. Rows 1–3, 4-6, and 7-9 correspond to Line, Dot, and Freeform types, respectively.}
\label{fig:supp_msp}
\end{figure}

Algorithms~\ref{alg:MSP-Line}, \ref{alg:MSP-Dot}, and \ref{alg:MSP-Freeform} describe the mask generation procedures for the Line, Dot, and Freeform types, respectively. Figure~\ref{fig:supp_msp} shows example primitive masks generated by these procedures.
On average, generating a single mask takes 0.54 ms (Line), 0.09 ms (Dot), and 115.23 ms (Freeform), based on 100 independent runs, demonstrating the efficiency of our procedural generation methods.

\begin{algorithm}[!ht]
\caption{Line Mask Generation Procedure}
\label{alg:MSP-Line}

\begin{algorithmic}[1]
\STATE $\mathbf{M} \gets$ initialize a mask of width $W$ and height $H$
\STATE $\mathbf{c}=(c_w,c_h) \sim (\text{U}(0, W),\ \text{U}(0, H))$\hfill \textit{\# center point}
\STATE $\theta \sim  \text{U}(0, \pi)$ \hfill \textit{\# radians angle}
\STATE $l \sim \text{U}(60, 200)$ \hfill \textit{\# length}
\STATE $s \sim \text{U}(20, 40)$ \hfill \textit{\# number of points}
\STATE $\mathbf{x} \in \mathbb{R}^s \leftarrow \text{linspace}(-l/2,\ l/2,\ s)$
\STATE $\mathbf{y} \in \mathbb{R}^s \leftarrow \mathbf{0}$
\IF{$\text{Bernoulli}(0.5)$}
    \STATE $\boldsymbol{\epsilon} \in \mathbb{R}^{s} \sim \text{N}(0,\ 14^2)$ \hfill \textit{\# Gaussian noise}
    \STATE $\mathbf{y} \leftarrow     \text{GaussianBlur}(\boldsymbol{\epsilon}, \ (1,5), \ 2)$ \\ \hfill    \textit{\# GaussianBlur(input, kernel size, std. dev.)}
\ENDIF
\STATE $\mathbf{t} \in \mathbb{R}^s \sim \text{U}(1,\ 7)$ \hfill \textit{\# thickness}
\FOR{$i = 1$ to $s - 1$}
    \STATE $\begin{bmatrix} x_i' \\ y_i' \end{bmatrix} \gets 
           \begin{bmatrix} \cos\theta & -\sin\theta \\ \sin\theta & \cos\theta \end{bmatrix}
           \begin{bmatrix} x_i \\ y_i \end{bmatrix} + \mathbf{c}$
    \STATE $\begin{bmatrix} x_{i+1}' \\ y_{i+1}' \end{bmatrix} \gets 
           \begin{bmatrix} \cos\theta & -\sin\theta \\ \sin\theta & \cos\theta \end{bmatrix}
           \begin{bmatrix} x_{i+1} \\ y_{i+1} \end{bmatrix} + \mathbf{c}$
    \STATE $\mathbf{M} \gets \mathbf{M} + \text{Line}((x_i', y_i'), (x_{i+1}', y_{i+1}'),\ t_i)$
\ENDFOR
\STATE $\mathbf{M} \leftarrow \text{Binarize}(\mathbf{M})$
\STATE \textbf{return} $\mathbf{M}$
\end{algorithmic}
\end{algorithm}

\begin{algorithm}[!ht]
\caption{Dot Mask Generation Procedure}
\label{alg:MSP-Dot}
\begin{algorithmic}[1]
\STATE $\mathbf{M} \gets$ initialize a mask of width $W$ and height $H$
\STATE $\mathbf{c}=(c_w,c_h) \sim (\text{U}(0, W),\ \text{U}(0, H))$\hfill \textit{\# center point}
\STATE $r \sim \text{U}(5, 35)$ \hfill \textit{\# radius}
\STATE $s \sim \text{U}(12, 30)$ \hfill \textit{\# number of points}
\STATE $\boldsymbol{\theta} \in \mathbb{R}^s \sim \text{U}(0,\ 2\pi)$ \hfill \textit{\# perimeter angles}
\STATE $\rho \sim \text{U}(0.6,\ 1.4)$ \hfill \textit{\# oval ratio}
\STATE $\beta \sim \text{U}(0.05,\ 0.35)$ \hfill \textit{\# radius jitter scale}
\STATE $u \sim \text{U}(0, 1)$
\STATE $r_i \sim 
\begin{cases}
\text{U}(-\beta r,\ \beta r) & \text{if } u \geq 0.66 \\
\text{N}(0,\ (\beta r)^2) & \text{if } u < 0.66
\end{cases}$, $\quad \forall i = 1, \dots, s$
\STATE $x_i \leftarrow c_x + (r + r_i)\cos \theta_i \cdot \rho,\quad \forall i = 1, \dots, s$
\STATE $y_i \leftarrow c_y + (r + r_i)\sin \theta_i,\quad \forall i = 1, \dots, s$
\STATE $\mathbf{M} \leftarrow \mathbf{M} + \text{Polygon}( \{(x_i, y_i)\}_{i=1}^{s})$
\IF{$\text{Bernoulli}(0.5)$}
    \STATE $k \sim \text{U}(\{3, 5, 7\})$
    \STATE $\mathbf{M} \leftarrow \text{GaussianBlur}(\mathbf{M},\ k,\ 0)$
\ENDIF
\STATE $\mathbf{M} \leftarrow \text{Binarize}(\mathbf{M})$
\STATE \textbf{return} $\mathbf{M}$
\end{algorithmic}
\end{algorithm}

\newpage

\begin{algorithm}[!ht]
\caption{Freeform Mask Generation Procedure}
\label{alg:MSP-Freeform}
\begin{algorithmic}[1]
\STATE $\mathbf{M} \gets$ initialize a mask of width $W$ and height $H$
\STATE $n_\text{step} \sim \text{U}(300,\ 18000)$ \hfill \textit{\# number of random walk steps}
\STATE $\sigma \sim \text{U}(2,\ 12)$ \hfill \textit{\# std. dev. for GaussianBlur}
\STATE $ (x_0,\ y_0) \sim (\text{U}(0, W),\ \text{U}(0, H))$ \hfill \textit{\# initial point}
\FOR{$i = 1$ to $n_\text{step}$}
    \STATE $(\Delta x_i,\ \Delta y_i) \sim (\text{U}(\{-1,\ 0,\ 1\}),\ \text{U}(\{-1,\ 0,\ 1\}))$
    \STATE $ (x_i,\ y_i) \leftarrow (x_{i-1},\ y_{i-1}) + (\Delta x_i,\ \Delta y_i)$
    \STATE $\mathbf{M}[y_i,\ x_i] \leftarrow 1$
\ENDFOR
\STATE $\mathbf{M} \leftarrow \text{GaussianBlur}(\mathbf{M},\ 2\cdot\sigma+1, \sigma)$
\IF{$\text{Bernoulli}(0.5)$}
    \STATE $\mathbf{M} \leftarrow \text{Dilate}(\mathbf{M},\ 3)$\hfill \textit{\# Dilate(kernel size)}
    \STATE $\mathbf{M} \leftarrow \text{Erode}(\mathbf{M},\ 3)$\hfill \textit{\# Erode(kernel size)}
\ENDIF
\STATE $\mathbf{M} \leftarrow \text{Binarize}(\mathbf{M})$
\STATE $\mathbf{M} \leftarrow \text{LargestComponent}(\mathbf{M})$ \hfill \textit{\# keep largest region}
\STATE \textbf{return} $\mathbf{M}$
\end{algorithmic}
\end{algorithm}

\onecolumn
\section{Two-Class Prompt Templates}\label{appendix:template}

In this work, we adopt the three state descriptor templates for each of the normal and anomaly states from WinCLIP~\cite{jeong2023winclip}, along with the 55 prompt templates from CLIPstyler~\cite{kwon2022clipstyler}. Therefore, each image is associated with 165 normal prompts and 165 anomaly prompts, \textit{i.e.}, $|\mathcal{T}^a|=|\mathcal{T}^n|=165$. 

\begin{figure*}[!ht]

\noindent
\begin{minipage}[t]{0.4\linewidth}
(a) \emph{state descriptor templates} - (normal state)

{\tt \small
\begin{itemize}
    \item s := "flawless [c]"
    \item s := "perfect [c]"
    \item s := "unblemished [c]"
\end{itemize}
}
\end{minipage}
\begin{minipage}[t]{0.4\linewidth}
(b) \emph{state descriptor templates} - (anomaly state)

{\tt \small
\begin{itemize}
    \item s := "[c] with [d] defect"
    \item s := "[c] with [d] flaw"
    \item s := "[c] with [d] damage"
\end{itemize}
}
\end{minipage}

\begin{minipage}[t]{0.33\linewidth}
\hfill
\end{minipage}

\vspace{5mm}
(c) \emph{prompt templates}

{\tt \small
\begin{multicols}{3}
\begin{itemize}
\item "a bad photo of a(n) [s]."
\item "a sculpture of a(n) [s]."
\item "a photo of the hard to see [s]."
\item "a low resolution photo of the [s]."
\item "a rendering of a(n) [s]."
\item "graffiti of a(n) [s]."
\item "a cropped photo of the [s]."
\item "a tattoo of a(n) [s]."
\item "the embroidered [s]."
\item "a photo of a hard to see [s]."
\item "a bright photo of a(n) [s]."
\item "a photo of a clean [s]."
\item "a photo of a dirty [s]."
\item "a dark photo of the [s]."
\item "a drawing of a(n) [s]."
\item "a photo of my [s]."
\item "the plastic [s]."
\item "a photo of the cool [s]."
\item "a close-up photo of a(n) [s]."
\item "a black and white photo of the [s]."
\item "a painting of the [s]."
\item "a painting of a(n) [s]."
\item "a pixelated photo of the [s]."
\item "a sculpture of the [s]."
\item "a bright photo of the [s]."
\item "a cropped photo of a(n) [s]."
\item "a plastic [s]."
\item "a photo of the dirty [s]."
\item "a blurry photo of the [s]."
\item "a photo of the [s]."
\item "a good photo of the [s]."
\item "a rendering of the [s]."
\item "a(n) [s] in a video game."
\item "a photo of one [s]."
\item "a doodle of a(n) [s]."
\item "a close-up photo of the [s]."
\item "a photo of a(n) [s]."
\item "the origami [s]."
\item "the [s] in a video game."
\item "a sketch of a(n) [s]."
\item "a doodle of the [s]."
\item "a origami [s]."
\item "a low resolution photo of a(n) [s]."
\item "the toy [s]."
\item "a rendition of the [s]."
\item "a photo of the clean [s]."
\item "a photo of a large [s]."
\item "a rendition of a(n) [s]."
\item "a photo of a nice [s]."
\item "a photo of a weird [s]."
\item "a blurry photo of a(n) [s]."
\item "a cartoon [s]."
\item "art of a [s]."
\item "a sketch of the [s]."
\item "a embroidered [s]."
\end{itemize}
\end{multicols}
}

\caption{Full list of two-class prompt templates. The category-defect pair (\texttt{[c]}, \texttt{[d]}) is inserted into the corresponding placeholders in the state descriptor templates to generate state descriptors for the normal and anomaly states. Each resulting state descriptor then replaces the placeholder \texttt{[s]} in the prompt templates to construct complete text prompts.}
\label{fig:comp_prompt}
\end{figure*}

\twocolumn

\section{Hyperparameter Settings}\label{appendix:hyperparam}

For style-guided mask generation, we set the decay coefficient $\alpha = 0.7$ and the maximum number of primitive masks $m_\text{max} = 5$. For text-driven localized anomaly generation, the number of patches $l$ was set to 64, with each patch sized at 128$\times$128. The distortion scale of the perspective transformation $\tau$ was set to 0.5. Since the input size to the image encoder $\mathcal{E}_I$ is 224$\times$224, we resized images or patches using bicubic interpolation before feeding them into $\mathcal{E}_I$. In the training objective, we adopted the following loss weights from CLIPstyler: $\lambda_{\text{gdir}} = 5 \times 10^2$, $\lambda_{\text{pdir}} = 9 \times 10^3$, $\lambda_{\text{tv}} = 2 \times 10^{-3}$, and $\lambda_\text{c} = 150$. We set the loss weight $\lambda_{\text{mclip}} = 10^3$. The content loss $L_c$ was computed using feature activations from the ``conv4\_2'' and ``conv5\_2'' layers of VGG-19, following \citet{gatys2016image}. The stylization network $\mathcal{F}$ was trained for 75 iterations using the Adam optimizer with a constant learning rate of $5 \times 10^{-4}$.

\section{Detailed Quantitative Results}\label{appendix:E}

\begin{table}[!ht]
    \centering
    \renewcommand{\arraystretch}{1.0}
    \setlength{\tabcolsep}{6pt}
    \resizebox{0.4\linewidth}{!}{%
    \begin{tabular}{l|cc}
        \toprule
        Category & IS & IC-L \\
        \midrule
        bottle      & 1.55 & 0.21 \\
        cable       & 1.53 & 0.39 \\
        capsule     & 1.96 & 0.20 \\
        carpet      & 1.28 & 0.26 \\
        grid        & 2.52 & 0.39 \\
        hazelnut    & 1.83 & 0.32 \\
        leather     & 2.94 & 0.45 \\
        metal nut   & 2.33 & 0.30 \\
        pill        & 1.65 & 0.24 \\
        screw       & 1.39 & 0.32 \\
        tile        & 2.92 & 0.52 \\
        toothbrush  & 1.46 & 0.17 \\
        transistor  & 1.57 & 0.29 \\
        wood        & 2.91 & 0.39 \\
        zipper      & 2.77 & 0.29 \\
        \midrule
        {Average} & {2.04} & {0.32} \\
        \bottomrule
    \end{tabular}
    }
    \caption{Per-category anomaly generation performance of AnoStyler on MVTec-AD.}
    \label{tab:mvtec_anostyler_only}
\end{table}

\begin{table}[!ht]
    \centering
    \renewcommand{\arraystretch}{1.0}
    \setlength{\tabcolsep}{6pt}
    \resizebox{0.4\linewidth}{!}{%
    \begin{tabular}{l|cc}
        \toprule
        Category & IS & IC-L \\
        \midrule
        candle       & 1.70 & 0.20 \\
        capsules     & 1.45 & 0.53 \\
        cashew       & 1.78 & 0.37 \\
        chewinggum   & 1.73 & 0.38 \\
        fryum        & 1.33 & 0.21 \\
        macaroni1    & 1.58 & 0.26 \\
        macaroni2    & 2.39 & 0.41 \\
        pcb1         & 1.28 & 0.35 \\
        pcb2         & 1.07 & 0.34 \\
        pcb3         & 1.14 & 0.23 \\
        pcb4         & 1.05 & 0.30 \\
        pipe fryum   & 2.14 & 0.24 \\
        \midrule
        {Average} & {1.55} & {0.32} \\
        \bottomrule
    \end{tabular}
    }
    \caption{Per-category anomaly generation performance of AnoStyler on VisA.}
    \label{tab:visa_anostyler_only}
\end{table}

\begin{table}[t]
\centering
\tabcolsep=4pt
\renewcommand{\arraystretch}{1.0}
\resizebox{\linewidth}{!}{%
\begin{tabular}{l|ccccccc}
\toprule
Category &
I-AUC & I-AP & I-F1 & P-AUC & P-AP & P-F1 & PRO \\
\midrule
bottle      & 98.8\scriptsize{±1.0} & 99.6\scriptsize{±0.3} & 97.8\scriptsize{±1.8} & 93.5\scriptsize{±1.2} & 70.0\scriptsize{±3.7} & 64.9\scriptsize{±2.8} & 84.6\scriptsize{±2.7} \\
cable       & 90.5\scriptsize{±1.0} & 94.0\scriptsize{±1.0} & 88.1\scriptsize{±0.2} & 89.9\scriptsize{±1.4} & 33.6\scriptsize{±6.1} & 41.7\scriptsize{±7.0} & 75.1\scriptsize{±0.0} \\
capsule     & 98.6\scriptsize{±0.7} & 99.7\scriptsize{±0.1} & 97.4\scriptsize{±0.2} & 94.9\scriptsize{±1.1} & 46.6\scriptsize{±4.3} & 47.0\scriptsize{±3.0} & 92.8\scriptsize{±0.1} \\
carpet      & 98.0\scriptsize{±0.6} & 99.5\scriptsize{±0.1} & 96.8\scriptsize{±0.6} & 97.8\scriptsize{±0.8} & 81.7\scriptsize{±1.7} & 75.7\scriptsize{±1.4} & 92.6\scriptsize{±0.9} \\
grid        & 100.0\scriptsize{±0.0} & 100.0\scriptsize{±0.0} & 100.0\scriptsize{±0.0} & 97.4\scriptsize{±0.0} & 41.4\scriptsize{±4.9} & 46.4\scriptsize{±2.3} & 94.0\scriptsize{±0.0} \\
hazelnut    & 99.3\scriptsize{±0.0} & 99.6\scriptsize{±0.0} & 98.1\scriptsize{±0.4} & 98.1\scriptsize{±0.1} & 74.0\scriptsize{±0.4} & 66.9\scriptsize{±0.6} & 96.1\scriptsize{±0.1} \\
leather     & 100.0\scriptsize{±0.0} & 100.0\scriptsize{±0.0} & 100.0\scriptsize{±0.0} & 99.8\scriptsize{±0.1} & 74.1\scriptsize{±1.0} & 67.6\scriptsize{±2.1} & 98.4\scriptsize{±0.4} \\
metal nut   & 94.3\scriptsize{±2.5} & 98.7\scriptsize{±0.5} & 94.3\scriptsize{±2.2} & 93.0\scriptsize{±1.6} & 71.8\scriptsize{±5.2} & 65.5\scriptsize{±2.9} & 80.1\scriptsize{±3.0} \\
pill        & 98.4\scriptsize{±0.2} & 99.7\scriptsize{±0.0} & 97.5\scriptsize{±0.3} & 99.4\scriptsize{±0.0} & 92.8\scriptsize{±0.5} & 85.6\scriptsize{±1.2} & 95.6\scriptsize{±0.1} \\
screw       & 96.2\scriptsize{±3.3} & 98.7\scriptsize{±1.0} & 95.0\scriptsize{±2.6} & 96.0\scriptsize{±0.9} & 31.0\scriptsize{±2.1} & 35.6\scriptsize{±4.0} & 88.1\scriptsize{±0.8} \\
tile        & 100.0\scriptsize{±0.0} & 100.0\scriptsize{±0.0} & 100.0\scriptsize{±0.0} & 99.5\scriptsize{±0.0} & 95.7\scriptsize{±0.3} & 88.7\scriptsize{±0.5} & 97.3\scriptsize{±0.1} \\
toothbrush  & 100.0\scriptsize{±0.0} & 100.0\scriptsize{±0.0} & 100.0\scriptsize{±0.0} & 97.8\scriptsize{±0.6} & 49.0\scriptsize{±2.4} & 54.6\scriptsize{±3.3} & 88.4\scriptsize{±1.1} \\
transistor  & 95.9\scriptsize{±1.5} & 95.7\scriptsize{±1.4} & 90.7\scriptsize{±2.6} & 64.4\scriptsize{±2.1} & 24.9\scriptsize{±5.7} & 27.1\scriptsize{±4.7} & 56.5\scriptsize{±1.1} \\
wood        & 99.8\scriptsize{±0.0} & 99.9\scriptsize{±0.0} & 99.2\scriptsize{±0.0} & 96.4\scriptsize{±0.2} & 79.1\scriptsize{±2.8} & 72.1\scriptsize{±3.2} & 92.4\scriptsize{±0.7} \\
zipper      & 100.0\scriptsize{±0.0} & 100.0\scriptsize{±0.0} & 100.0\scriptsize{±0.0} & 98.3\scriptsize{±0.4} & 77.8\scriptsize{±2.8} & 71.8\scriptsize{±2.5} & 92.8\scriptsize{±1.1} \\
\midrule
{Average}   & {98.0}\scriptsize{±0.3} & {99.0}\scriptsize{±0.2} & {97.0}\scriptsize{±0.4} & {94.4}\scriptsize{±0.5} & {62.9}\scriptsize{±1.1} & {60.7}\scriptsize{±1.1} & {88.3}\scriptsize{±0.1} \\
\bottomrule
\end{tabular}%
}
\caption{Per-category anomaly detection performance of AnoStyler on MVTec-AD (mean$\pm$standard deviation over 5 runs).}
\label{tab:mvtec_per_category}
\end{table}

\begin{table}[t]
\centering
\tabcolsep=4pt
\renewcommand{\arraystretch}{1.0}
\resizebox{\linewidth}{!}{%
\begin{tabular}{l|ccccccc}
\toprule
Category &
I-AUC & I-AP & I-F1 & P-AUC & P-AP & P-F1 & PRO \\
\midrule
candle       & 91.4\scriptsize{±2.0} & 92.8\scriptsize{±1.3} & 85.7\scriptsize{±1.8} & 90.7\scriptsize{±0.9} & 23.1\scriptsize{±1.1} & 32.4\scriptsize{±0.4} & 84.3\scriptsize{±0.8} \\
capsules     & 93.1\scriptsize{±2.0} & 96.2\scriptsize{±1.1} & 90.2\scriptsize{±2.5} & 92.7\scriptsize{±2.5} & 26.4\scriptsize{±5.4} & 34.5\scriptsize{±4.8} & 86.5\scriptsize{±3.4} \\
cashew       & 95.8\scriptsize{±0.0} & 97.4\scriptsize{±0.4} & 94.6\scriptsize{±1.8} & 92.9\scriptsize{±0.5} & 18.7\scriptsize{±10.6} & 23.3\scriptsize{±8.6} & 88.4\scriptsize{±10.3} \\
chewinggum   & 98.3\scriptsize{±0.1} & 99.3\scriptsize{±0.0} & 96.6\scriptsize{±0.3} & 98.9\scriptsize{±0.0} & 75.3\scriptsize{±9.1} & 67.0\scriptsize{±10.1} & 84.2\scriptsize{±2.1} \\
fryum        & 87.9\scriptsize{±2.5} & 94.2\scriptsize{±1.7} & 87.8\scriptsize{±1.7} & 93.2\scriptsize{±2.5} & 30.6\scriptsize{±3.7} & 34.6\scriptsize{±2.5} & 86.5\scriptsize{±6.6} \\
macaroni1    & 97.0\scriptsize{±0.3} & 97.6\scriptsize{±0.3} & 92.2\scriptsize{±0.5} & 97.9\scriptsize{±0.6} & 24.1\scriptsize{±5.8} & 31.7\scriptsize{±5.5} & 92.2\scriptsize{±1.5} \\
macaroni2    & 89.5\scriptsize{±2.1} & 91.5\scriptsize{±2.2} & 83.3\scriptsize{±3.4} & 96.3\scriptsize{±1.1} & 15.3\scriptsize{±4.1} & 25.2\scriptsize{±4.5} & 89.7\scriptsize{±0.3} \\
pcb1         & 88.2\scriptsize{±3.8} & 88.4\scriptsize{±3.8} & 82.7\scriptsize{±3.4} & 93.4\scriptsize{±0.7} & 46.4\scriptsize{±15.0} & 49.4\scriptsize{±17.0} & 62.2\scriptsize{±3.3} \\
pcb2         & 95.5\scriptsize{±1.0} & 95.5\scriptsize{±0.8} & 90.9\scriptsize{±2.2} & 89.7\scriptsize{±1.0} & 17.1\scriptsize{±9.0} & 22.5\scriptsize{±7.0} & 72.5\scriptsize{±2.7} \\
pcb3         & 95.7\scriptsize{±1.1} & 96.3\scriptsize{±1.0} & 89.7\scriptsize{±1.4} & 89.3\scriptsize{±2.3} & 26.3\scriptsize{±1.9} & 34.0\scriptsize{±2.0} & 88.5\scriptsize{±0.8} \\
pcb4         & 97.3\scriptsize{±0.0} & 96.0\scriptsize{±0.4} & 93.7\scriptsize{±0.2} & 94.4\scriptsize{±0.5} & 32.3\scriptsize{±1.7} & 37.4\scriptsize{±1.5} & 84.0\scriptsize{±0.6} \\
pipe fryum   & 96.8\scriptsize{±1.4} & 98.4\scriptsize{±0.7} & 93.9\scriptsize{±1.6} & 96.0\scriptsize{±1.6} & 40.1\scriptsize{±2.0} & 45.2\scriptsize{±1.1} & 93.2\scriptsize{±1.5} \\
\midrule
{Average}   & {93.9}\scriptsize{±1.0} & {95.3}\scriptsize{±0.7} & {90.1}\scriptsize{±0.9} & {93.8}\scriptsize{±0.5} & {31.4}\scriptsize{±1.4} & {36.4}\scriptsize{±1.5} & {84.3}\scriptsize{±0.6} \\
\bottomrule
\end{tabular}%
}
\caption{Per-category anomaly detection performance of AnoStyler on VisA (mean$\pm$standard deviation over 5 runs).}
\label{tab:visa_per_category}
\end{table}

\begin{table}[!t]
\centering
\tabcolsep=5pt
\renewcommand{\arraystretch}{1.0}
\resizebox{\linewidth}{!}{  
\begin{tabular}{c|ccc|cc|cc}
\toprule
\#
 & \shortstack{$\mathcal{L}_\text{gdir}$} & \shortstack{$\mathcal{L}_\text{pdir}$} & \shortstack{$\mathcal{L}_\text{mclip}$} & IS  & IC-L  & I-AUC & P-AUC  \\
\midrule
(a) &  &  &  & 1.40 & 0.24 & 89.3 & 85.3 \\
(b) & \checkmark &  &  & 1.48 & 0.27 & 90.2 & 89.5 \\
(c) & \checkmark & \checkmark &  & 1.51 & 0.29 & 91.6 & 91.7 \\
(d) & \checkmark & \checkmark & \checkmark & \textbf{1.55} & \textbf{0.32} & \textbf{93.9} & \textbf{93.8} \\
\bottomrule
\end{tabular}}
\caption{Ablation study on the effect of the proposed loss components in AnoStyler on the VisA dataset.}
\label{tab:loss_ablation_visa}
\end{table}

\onecolumn
\section{Detailed Qualitative Results}\label{appendix:F}

\begin{figure*}[!ht]
\centering
\includegraphics[width=0.9\linewidth]{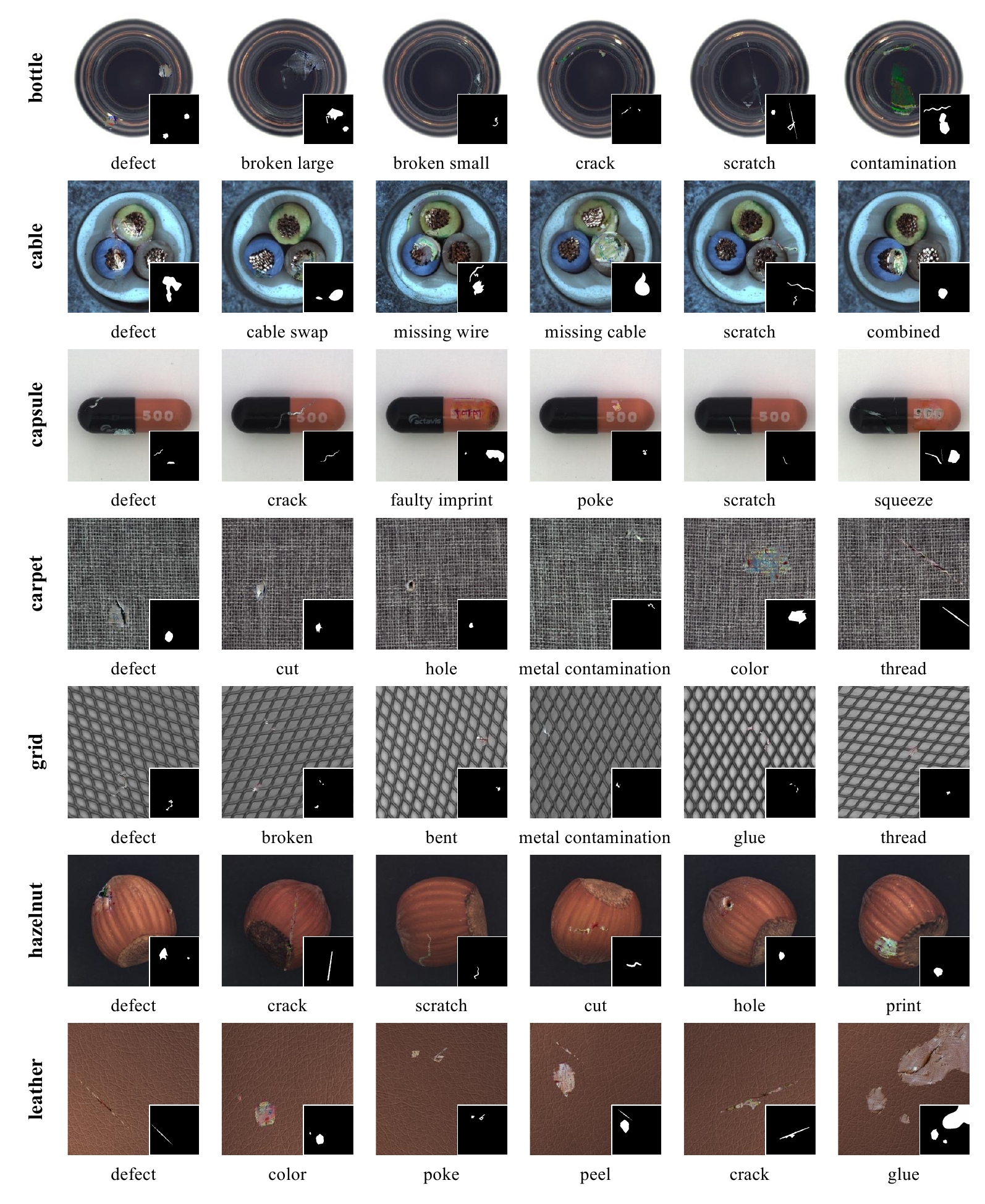}
\caption{Generated anomaly images for various defect types in MVTec-AD. Each image is shown with its corresponding anomaly mask (bottom-right) (continued).}
\label{fig:supp_mvtec1}
\end{figure*}

\begin{figure*}[!ht]
\centering
\ContinuedFloat
\includegraphics[width=0.9\linewidth]{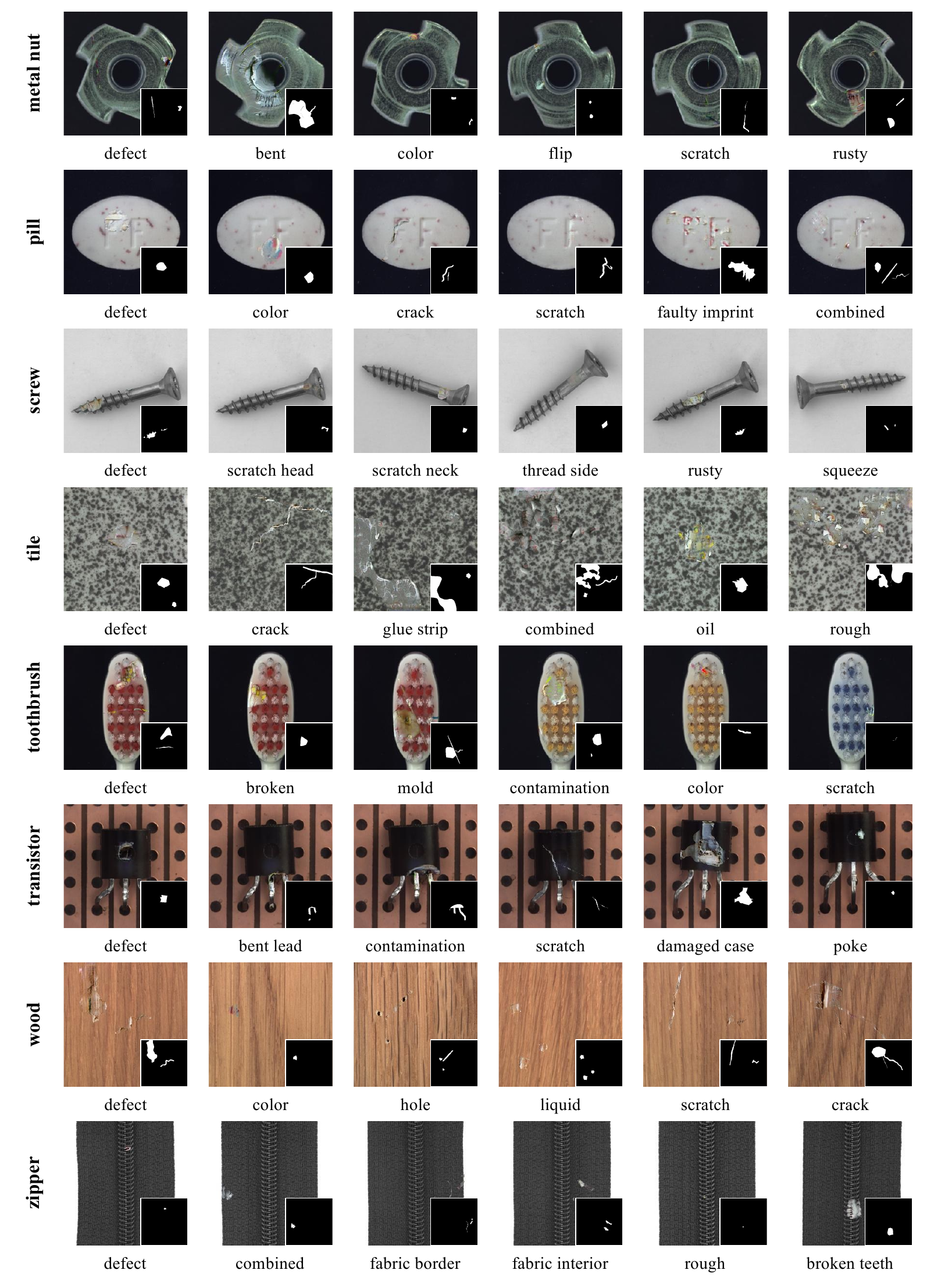}
\caption{Generated anomaly images for various defect types in MVTec-AD. Each image is shown with its corresponding anomaly mask (bottom-right).}
\label{fig:supp_mvtec2}
\end{figure*}

\begin{figure*}[!ht]
\centering
\includegraphics[width=0.9\linewidth]{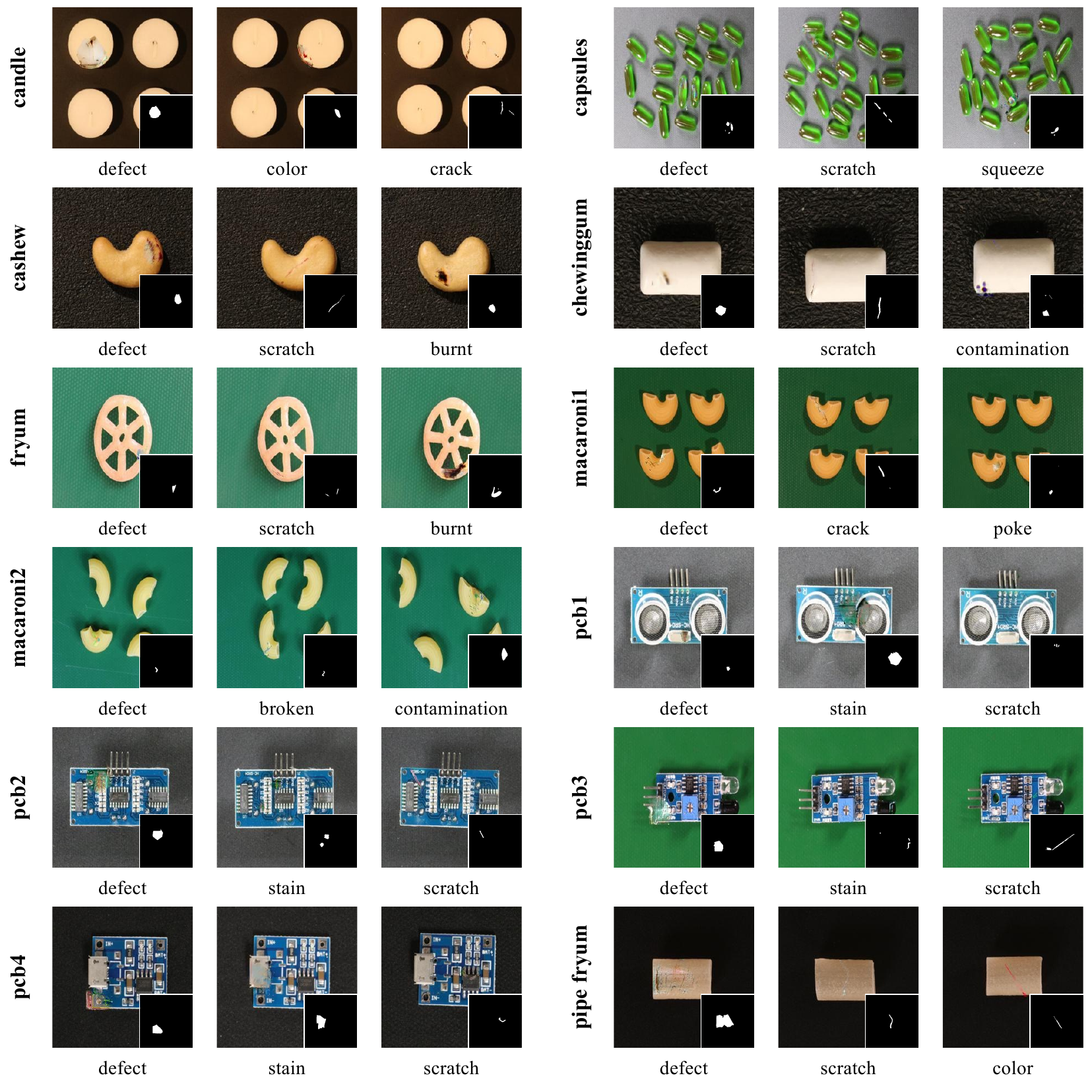}
\caption{Generated anomaly images for various defect types in VisA. Each image is shown with its corresponding anomaly mask (bottom-right).}
\label{fig:supp_visa}
\end{figure*}

\clearpage
\FloatBarrier
\section{Statistical Significance Analysis}\label{appendix:G}

\begin{figure*}[!ht]
\centering
\includegraphics[width=\linewidth]{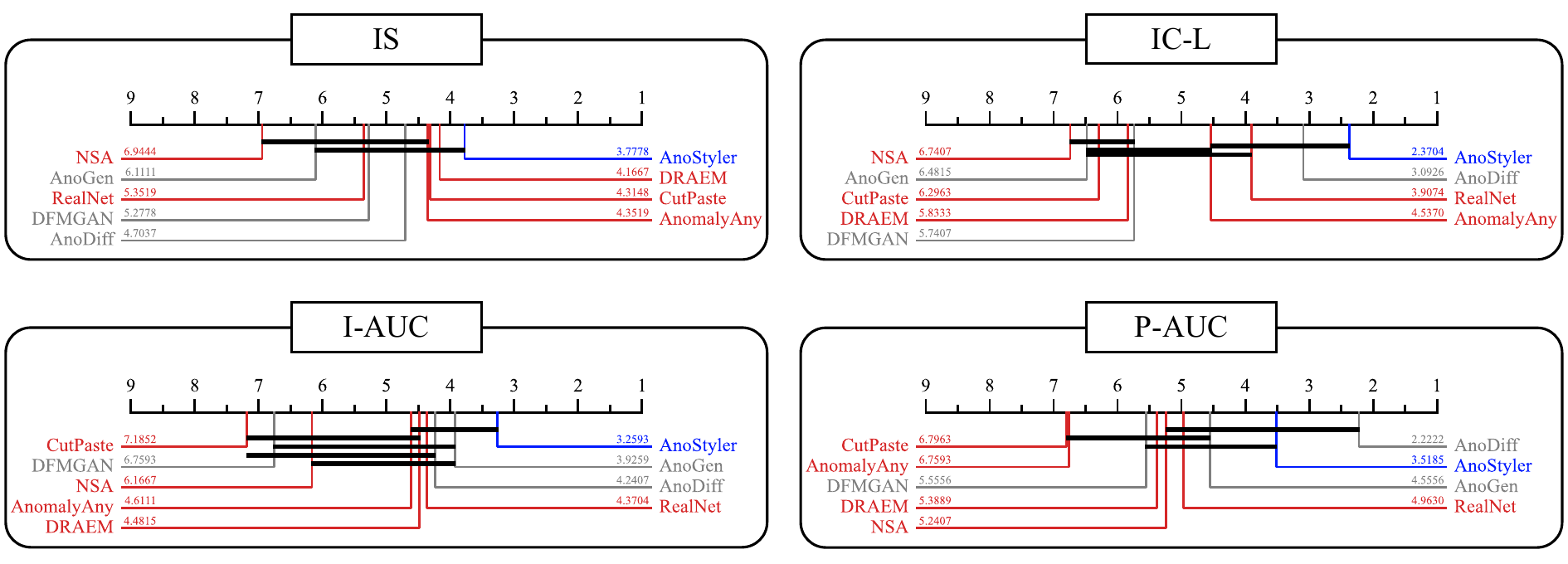}

\caption{Critical difference diagrams of 9 methods evaluated on MVTec-AD and VisA by comparing their ranks for each category using the IS, IC-L, I-AUC, and P-AUC metrics.{AnoStyler} is shown in blue, zero-shot baselines in red, and few-shot baselines in gray.}
\label{fig:supp_G}
\end{figure*}

We applied the Friedman test to each metric, and the null hypothesis was rejected in all cases (IS: $p=1.79\times10^{-4}$, IC-L: $p=8.64\times10^{-13}$, I-AUC: $p=2.61\times10^{-9}$, P-AUC: $p=2.48\times10^{-10}$), indicating significant differences among methods. Pairwise post-hoc Wilcoxon signed-rank tests with Holm's correction were then performed at a significance level of $0.05$, and the results are visualized in Figure~\ref{fig:supp_G} as critical difference diagrams.
For the generation metrics, {AnoStyler} achieved the best average rank in both IS and IC-L. Specifically, {AnoStyler} significantly outperformed NSA in IS, and outperformed DFMGAN, NSA, DRAEM, CutPaste, and AnoGen in IC-L. While other methods exhibited inconsistent rankings between IS and IC-L, {AnoStyler} consistently achieved top performance in both metrics, indicating that it generates anomaly images with both high visual quality and high diversity.
For the detection metrics, {AnoStyler} ranked first in I-AUC and second in P-AUC. It significantly outperformed CutPaste, NSA, and DFMGAN in I-AUC, and outperformed CutPaste and AnomalyAny in P-AUC. Although the differences with AnoGen and AnoDiff were not statistically significant, {AnoStyler} achieved comparable performance to these few-shot methods.
Overall, these results indicate that {AnoStyler} outperforms most zero-shot baselines (CutPaste, DRAEM, NSA, RealNet, AnomalyAny) and also surpasses a few-shot method (DFMGAN) while remaining competitive with AnoGen and AnoDiff.

%% file: aaai2026.bib
@inproceedings{gatys2016image,
  author    = {Leon A. Gatys and Alexander S. Ecker and Matthias Bethge},
  title     = {Image Style Transfer Using Convolutional Neural Networks},
  booktitle = {Proceedings of the IEEE Conference on Computer Vision and Pattern Recognition},
  year      = {2016},
  pages     = {2414--2423},
}

@article{gal2021stylegannada,
  author    = {Rinon Gal and Yuval Alaluf and Yuval Atzmon and Amit H. Bermano and Gal Chechik and Daniel Cohen-Or},
  title     = {StyleGAN-NADA: CLIP-Guided Domain Adaptation of Image Generators},
  journal = {ACM Transactions on Graphics},
  volume={41},
  number={4},
  year = {2022},
  pages={141}
}

@inproceedings{li2017universal,
  author    = {Yijun Li and Chen Fang and Jimei Yang and Zhaowen Wang and Xin Lu and Ming-Hsuan Yang},
  title     = {Universal Style Transfer via Feature Transforms},
  booktitle = {Advances in Neural Information Processing Systems},
  year      = {2017},
  pages     = {386--396}
}

@inproceedings{li2018closed,
  author    = {Yijun Li and Ming-Yu Liu and Xueting Li and Ming-Hsuan Yang and Jan Kautz},
  title     = {A Closed-Form Solution to Photorealistic Image Stylization},
  booktitle = {Proceedings of the European Conference on Computer Vision},
  year      = {2018},
  pages     = {453--468}
}

@inproceedings{huang2017adain,
  author    = {Xun Huang and Serge Belongie},
  title     = {Arbitrary Style Transfer in Real-Time with Adaptive Instance Normalization},
  booktitle = {Proceedings of the IEEE International Conference on Computer Vision},
  year      = {2017},
  pages     = {1501--1510}
}

@inproceedings{radford2021clip,
  author    = {Alec Radford and Jong Wook Kim and Chris Hallacy and Aditya Ramesh and Gabriel Goh and Sandhini Agarwal and Girish Sastry and Amanda Askell and Pamela Mishkin and Jack Clark and Gretchen Krueger and Ilya Sutskever},
  title     = {Learning Transferable Visual Models from Natural Language Supervision},
  booktitle = {Proceedings of the International Conference on Machine Learning},
  year      = {2021},
  pages     = {8748--8763}
}

@inproceedings{patashnik2021styleclip,
  author    = {Or Patashnik and Zongze Wu and Eli Shechtman and Daniel Cohen-Or and Dani Lischinski},
  title     = {StyleCLIP: Text-Driven Manipulation of StyleGAN Imagery},
  booktitle = {Proceedings of the IEEE/CVF International Conference on Computer Vision},
  year      = {2021},
  pages     = {2085--2094}
}

@inproceedings{karras2019stylegan,
  author    = {Tero Karras and Samuli Laine and Timo Aila},
  title     = {A Style-Based Generator Architecture for Generative Adversarial Networks},
  booktitle = {Proceedings of the IEEE/CVF Conference on Computer Vision and Pattern Recognition},
  year      = {2019},
  pages     = {4401--4410}
}

@inproceedings{kwon2022clipstyler,
  author    = {Hyungjin Kwon and Dahun Kim and Yunjey Choi and Jonghyun Kim and Jung-Woo Ha},
  title     = {CLIPstyler: Image Style Transfer with a Single Text Condition},
  booktitle = {Proceedings of the IEEE/CVF Conference on Computer Vision and Pattern Recognition},
  year      = {2022},
  pages     = {18062--18071}
}

@inproceedings{ho2020ddpm,
  author    = {Jonathan Ho and Ajay Jain and Pieter Abbeel},
  title     = {Denoising Diffusion Probabilistic Models},
  booktitle = {Advances in Neural Information Processing Systems},
  year      = {2020},
  pages     = {6840--6851}
}

@inproceedings{bergmann2019mvtec,
  author    = {Paul Bergmann and Michael Fauser and David Sattlegger and Carsten Steger},
  title     = {MVTec AD: A Comprehensive Real-World Dataset for Unsupervised Anomaly Detection},
  booktitle = {Proceedings of the IEEE/CVF Conference on Computer Vision and Pattern Recognition},
  year      = {2019},
  pages     = {9592--9600}
}

@inproceedings{zou2022spot,
  author    = {Yang Zou and Jongheon Jeong and Latha Pemula and Dongqing Zhang and Onkar Dabeer},
  title     = {SPot-the-Difference Self-Supervised Pre-training for Anomaly Detection and Segmentation},
  booktitle = {Proceedings of the European Conference on Computer Vision},
  year      = {2022},
  pages     = {392--408}
}

@inproceedings{ojha2021fewshot,
  author    = {Utkarsh Ojha and Yijun Li and Jingwan Lu and Alexei A. Efros and Yong Jae Lee and Eli Shechtman and Richard Zhang},
  title     = {Few-Shot Image Generation via Cross-Domain Correspondence},
  booktitle = {Proceedings of the IEEE/CVF Conference on Computer Vision and Pattern Recognition},
  year      = {2021},
  pages     = {10743--10752}
}

@inproceedings{jeong2023winclip,
  author    = {Jeong, Jongheon and Zou, Yang and Kim, Taewan and Zhang, Dongqing and Ravichandran, Avinash and Dabeer, Onkar},
  title     = {WinCLIP: Zero-/Few-Shot Anomaly Classification and Segmentation},
  booktitle = {Proceedings of the IEEE/CVF Conference on Computer Vision and Pattern Recognition},
  year      = {2023},
  pages     = {19606--19616}
}

@inproceedings{batzner2024efficientad,
  author    = {Batzner, Kilian and Heckler, Lars and K{\"o}nig, Rebecca},
  title     = {EfficientAD: Accurate Visual Anomaly Detection at Millisecond-Level Latencies},
  booktitle = {Proceedings of the IEEE/CVF Winter Conference on Applications of Computer Vision},
  year      = {2024},
  pages     = {128--138},
}

@inproceedings{roth2022total,
  author    = {Karsten Roth and Latha Pemula and Joaquin Zepeda and Bernhard Sch{\"o}lkopf and Thomas Brox and Peter Gehler},
  title     = {Towards Total Recall in Industrial Anomaly Detection},
  booktitle = {Proceedings of the IEEE/CVF Conference on Computer Vision and Pattern Recognition},
  year      = {2022},
  pages     = {14318--14328},
}

@inproceedings{defard2021padim,
  author    = {Thomas Defard and Aleksandr Setkov and Angelique Loesch and Romaric Audigier},
  title     = {PaDiM: a Patch Distribution Modeling Framework for Anomaly Detection and Localization},
  booktitle = {Proceedings of the International Conference on Pattern Recognition},
  year      = {2021},
  pages     = {475--489}
}

@inproceedings{hu2024anomalydiffusion,
  author    = {Teng Hu and Jiangning Zhang and Ran Yi and Yuzhen Du and Xu Chen and Liang Liu and Yabiao Wang and Chengjie Wang},
  title     = {AnomalyDiffusion: Few-Shot Anomaly Image Generation with Diffusion Model},
  booktitle = {Proceedings of the AAAI Conference on Artificial Intelligence},
  year      = {2024},
  pages     = {8526--8534}
}

@inproceedings{duan2023defect,
  author    = {Yuxuan Duan and Yan Hong and Li Niu and Liqing Zhang},
  title     = {Few-Shot Defect Image Generation via Defect-Aware Feature Manipulation},
  booktitle = {Proceedings of the AAAI Conference on Artificial Intelligence},
  year      = {2023},
  pages     = {571--578}
}

@inproceedings{gui2024anogen,
  author    = {Guan Gui and Bin-Bin Gao and Jun Liu and Chengjie Wang and Yunsheng Wu},
  title     = {Few-Shot Anomaly-Driven Generation for Anomaly Classification and Segmentation},
  booktitle = {Proceedings of the European Conference on Computer Vision},
  year      = {2024},
  pages     = {210--226}
}

@inproceedings{lin2021few,
  author    = {Dongyun Lin and Yanpeng Cao and Wenbing Zhu and Yiqun Li},
  title     = {Few-Shot Defect Segmentation Leveraging Abundant Normal Training Samples Through Normal Background Regularization and Crop-and-Paste Operation},
  booktitle = {Proceedings of the IEEE International Conference on Multimedia and Expo},
  year      = {2021}}

@inproceedings{li2021cutpaste,
  author    = {Chun-Liang Li and Kihyuk Sohn and Jinsung Yoon and Tomas Pfister},
  title     = {CutPaste: Self-Supervised Learning for Anomaly Detection and Localization},
  booktitle = {Proceedings of the IEEE/CVF Conference on Computer Vision and Pattern Recognition},
  year      = {2021},
  pages     = {9664--9674}
}

@inproceedings{schluter2022natural,
  author    = {Hannah M. Schlüter and Jeremy Tan and Benjamin Hou and Bernhard Kainz},
  title     = {Natural Synthetic Anomalies for Self-Supervised Anomaly Detection and Localization},
  booktitle = {Proceedings of the European Conference on Computer Vision},
  year      = {2022},
  pages     = {474--489}
}

@inproceedings{zavrtanik2021draem,
  author    = {Vitjan Zavrtanik and Matej Kristan and Danijel Sko{\v{c}}aj},
  title     = {DRAEM -- A Discriminatively Trained Reconstruction Embedding for Surface Anomaly Detection},
  booktitle = {Proceedings of the IEEE/CVF International Conference on Computer Vision},
  year      = {2021},
  pages     = {8330--8339}
}

@inproceedings{goodfellow2014generative,
  author    = {Ian J. Goodfellow and Jean Pouget-Abadie and Mehdi Mirza and Bing Xu and David Warde-Farley and Sherjil Ozair and Aaron Courville and Yoshua Bengio},
  title     = {Generative Adversarial Nets},
  booktitle = {Advances in Neural Information Processing Systems},
  year      = {2014},
  pages     = {2672--2680}
}

@inproceedings{zhang2024realnet,
  author    = {Ximiao Zhang and Min Xu and Xiuzhuang Zhou},
  title     = {RealNet: A Feature Selection Network with Realistic Synthetic Anomaly for Anomaly Detection},
  booktitle = {Proceedings of the IEEE/CVF Conference on Computer Vision and Pattern Recognition},
  year      = {2024},
  pages     = {16699--16709}
}

@inproceedings{sun2025unseen,
  author    = {Han Sun and Yunkang Cao and Hao Dong and Olga Fink},
  title     = {Unseen Visual Anomaly Generation},
  booktitle = {Proceedings of the IEEE/CVF Conference on Computer Vision and Pattern Recognition},
  year      = {2025},
  pages     = {25508--25517}
}

@inproceedings{singh2024least,
  author    = {Silky Singh and Surgan Jandial and Simra Shahid and Abhinav Java},
  title     = {LEAST: ``Local'' text-conditioned image style transfer},
  booktitle = {Proceedings of the CVPR Workshop on AI for Content Creation},
  year      = {2024}
}

@inproceedings{kirillov2023segment,
  author    = {Alexander Kirillov and Eric Mintun and Nikhila Ravi and Hanzi Mao and Chloe Rolland and Laura Gustafson and Tete Xiao and Spencer Whitehead and Alexander C. Berg and Wan-Yen Lo and Piotr Dollár and Ross Girshick},
  title     = {Segment Anything},
  booktitle = {Proceedings of the IEEE/CVF Conference on Computer Vision and Pattern Recognition},
  year      = {2023},
  pages     = {4015--4026}
}

@inproceedings{chen2024soulstyler,
  author    = {Junhao Chen and Peng Rong and Jingbo Sun and Chao Li and Xiang Li and Hongwu Lv},
  title     = {Soulstyler: Using Large Language Model to Guide Image Style Transfer for Target Object},
  booktitle = {Proceedings of the IEEE International Conference on Acoustics, Speech and Signal Processing},
  year      = {2024},
  pages     = {4015--4020}
}

@article{rudin1992nonlinear,
  author    = {Leonid I. Rudin and Stanley Osher and Emad Fatemi},
  title     = {Nonlinear Total Variation Based Noise Removal Algorithms},
  journal   = {Physica D: Nonlinear Phenomena},
  volume    = {60},
  number    = {1--4},
  pages     = {259--268},
  year      = {1992},
  publisher = {Elsevier}
}

@inproceedings{salimans2016improved,
  author    = {Tim Salimans and Ian Goodfellow and Wojciech Zaremba and Vicki Cheung and Alec Radford and Xi Chen},
  title     = {Improved Techniques for Training GANs},
  booktitle = {Advances in Neural Information Processing Systems},
  year      = {2016},
  pages     = {2234--2242}
}

@article{li2025survey,
  author    = {Zhuo Li and Yuhao Yan and Xiangheng Wang and Yifei Ge and Lin Meng},
  title     = {A Survey of Deep Learning for Industrial Visual Anomaly Detection},
  journal   = {Artificial Intelligence Review},
  volume    = {58},
  number    = {279},
  year      = {2025}
}

@inproceedings{hyun2024reconpatch,
  author       = {Jeeho Hyun and Sangyun Kim and Giyoung Jeon and Seung Hwan Kim and Kyunghoon Bae and Byung Jun Kang},
  title        = {ReConPatch: Contrastive Patch Representation Learning for Industrial Anomaly Detection},
  booktitle    = {Proceedings of the IEEE/CVF Winter Conference on Applications of Computer Vision},
  year         = {2024},
  pages        = {2052--2061}
}

@inproceedings{zhang2021defectgan,
  author       = {Gongjie Zhang and Kaiwen Cui and Tzu‑Yi Hung and Shijian Lu},
  title        = {Defect‑GAN: High‑Fidelity Defect Synthesis for Automated Defect Inspection},
  booktitle    = {Proceedings of the IEEE/CVF Winter Conference on Applications of Computer Vision},
  year         = {2021},
  pages        = {2524--2534}
}

@inproceedings{ganugula2023mosaic,
  author    = {Prajwal Ganugula and Y.\ S.\ S.\ S.\ Santosh Kumar and N.\ K.\ Sagar Reddy and Prabhath Chellingi and Avinash Thakur and Neeraj Kasera and C.\ Shyam Anand},
  title     = {MOSAIC: Multi‑Object Segmented Arbitrary Stylization Using CLIP},
  booktitle = {Proceedings of the IEEE/CVF International Conference on Computer Vision Workshops},
  year      = {2023},
  pages     = {892--903}
}

@inproceedings{kamra2023semcs,
  author    = {Chanda Grover Kamra and Indra Deep Mastan and Debayan Gupta},
  title     = {SEM‑CS: Semantic CLIPStyler for Text‑Based Image Style Transfer},
  booktitle = {Proceedings of the IEEE International Conference on Image Processing},
  year      = {2023},
  pages     = {395--399}
}

@article{cui2023survey,
  author    = {Yajie Cui and Zhaoxiang Liu and Shiguo Lian},
  title     = {A Survey on Unsupervised Anomaly Detection Algorithms for Industrial Images},
  journal   = {IEEE Access},
  year      = {2023},
  volume    = {11},
  pages     = {55297--55315}
}

@inproceedings{cao2023anomaldistshift,
  author    = {Tri Thien Cao and Jiawen Zhu and Guansong Pang},
  title     = {Anomaly Detection Under Distribution Shift},
  booktitle = {Proceedings of the IEEE/CVF International Conference on Computer Vision},
  year      = {2023},
  pages     = {6511--6523}
}

@inproceedings{wu2025dfm,
  author    = {Wu, Haoning and Xu, Qianli and Li, Zihan and Liang, Song and Chen, Liang and Zhao, Hang and Lin, Dahua and Xu, Wei and Wei, Yunchao},
  title     = {DFM: Differentiable Feature Matching for Anomaly Detection},
  booktitle = {Proceedings of the IEEE/CVF Conference on Computer Vision and Pattern Recognition},
  year      = {2025},
  pages     = {15224--15233}
}

@inproceedings{fang2025ckaad,
  author    = {Fang, Qingqing and Su, Qinliang and Lv, Wenxi and Xu, Wenchao and Yu, Jianxing},
  title     = {Boosting Fine‑Grained Visual Anomaly Detection with Coarse‑Knowledge‑Aware Adversarial Learning},
  booktitle = {Proceedings of the AAAI Conference on Artificial Intelligence},
  year      = {2025},
  pages     = {7943--7951}
}
